RESEARCH PAPER

# A generative adversarial network optimization method for damage detection and digital twinning by deep AI fault learning: Z24 Bridge structural health monitoring benchmark validation

**Marios Impraimakis**[1] · **Evangelia Nektaria Palkanoglou**[2]

Received: 14 June 2025 / Revised: 25 September 2025 / Accepted: 29 September 2025

**Abstract**

The optimization-based damage detection and damage state digital twinning capabilities are examined herein of a novel conditional-labeled generative adversarial network methodology. The framework outperforms current approaches for fault anomaly detection as no prior information is required for the health state of the system: a topic of high significance for real-world applications. Specifically, current artificial intelligence-based digital twinning approaches suffer from the uncertainty related to obtaining poor predictions when a low number of measurements is available, physics knowledge is missing, or when the damage state is unknown. To this end, an unsupervised framework is examined and validated rigorously on the benchmark structural health monitoring measurements of Z24 Bridge: a post-tensioned concrete highway bridge in Switzerland, as a part of a full-scale monitoring and controlled damage experiment. In implementing the approach, firstly, different same damage-level measurements are used as inputs, while the model is forced to converge conditionally to two different damage states. Secondly, the process is repeated for a different group of measurements. Finally, the convergence scores are compared to identify which one belongs to a different damage state. The network optimization process for both healthy-to-healthy and damage-to-healthy input data creates, simultaneously, measurements for digital twinning purposes at different damage states, capable of pattern recognition and machine learning data generation. In contrast to conventional supervised methods, the proposed conditional-labeled generative adversarial network enables both unsupervised damage detection and generation of damage state measurements. Further to this process, a support vector machine classifier and a principal component analysis procedure is developed to assess the generated and real measurements of each damage category, serving as a secondary new dynamics learning indicator in damage scenarios. Importantly, the approach is shown to capture accurately damage over healthy measurements, providing a powerful tool for vibration-based system-level monitoring and scalable infrastructure resilience.

**Keywords** Structural health monitoring optimization · Generative adversarial networks data augmentation · Digital twin pattern recognition · Deep learning damage detection · Unsupervised nondestructive testing evaluation · (1D) one-dimensional convolutional neural networks

## 1 Introduction

Damage detection in digital twinning processes is not a trivial problem as measurement anomalies occur for many more reasons than damage. This is especially true in global large-scale structural system identification, which directly affects the structural safety and resilience (Bruneau et al. 2003; Cimellaro et al. 2010). To this end, structural health monitoring (Farrar and Worden 2007) with or without modern artificial intelligence tools is successfully implemented (Cha et al. 2024; Azimi and Pekcan 2020; Bao et al. 2019; Azimi et al. 2020; Seventekidis et al. 2020; Dang et al.

✉ Marios Impraimakis
mi595@bath.ac.uk

1 Department of Mechanical Engineering, University of Bath, BA2 7AY Bath, UK

2 Wolfson School of Mechanical, Electrical and Manufacturing Engineering, Loughborough University, LE11 3TU Loughborough, UK

2020; Malekloo et al. 2022). These methodologies provide practical tools for assessing the condition of monitored systems through continuous and often real-time measurements (Kijewski-Correa et al. 2013; Kim and Feng 2007; Masri et al. 2004; Rainieri et al. 2011; Kaya and Safak 2015; Impraimakis and Smyth 2022; Impraimakis and Smyth 2022c; Impraimakis 2024b). The processes often enable early damage detection (Roveri et al. 2025; Bernagozzi et al. 2022, maintenance optimization (Andriotis and Papakonstantinou 2019; Arcieri et al. 2023; Morato et al. 2023), and life cycle extension (Torti et al. 2022; Okasha et al. 2010; Smarsly et al. 2013; Bhattacharya et al. 2025). Real-time estimation is especially crucial for realistic digital twins evolving in time (Chua et al. 2025; Razmarashooli et al. 2025). However, the deployment of effective structural health monitoring frameworks remains challenging in scarce measurement cases due to the complexity of sensor inaccuracies and the structural dynamics changes under variable environmental and loading conditions (Velde et al. 2025; Deraemaeker and Worden 2018; Sohn 2007; Keshmiry et al. 2023; Erazo et al. 2019; Deraemaeker et al. 2008; Catbas et al. 2008).

Recent advances in artificial intelligence, particularly in deep generative models, offer promising pathways to overcome limitations related to these complexities for creating better digital twins (Wagg et al. 2020; Thelen et al. 2022). Generative adversarial networks, introduced by (Goodfellow et al. 2014) have shown remarkable success in generating synthetic data that mimic real-world distributions in fields such as image synthesis, audio reconstruction, and medical diagnostics. Yet, their application in structural dynamics and structural health monitoring remains largely unexplored, especially for time series vibration data in real-world damage scenarios. In structural health monitoring applications, (Tsialiamanis et al. 2022) introduced a machine learning scheme for nonlinear modal analysis applications, while (Maeda et al. 2021) used generative models to generate road infrastructure fault images that cannot be distinguished from a real one. Furthermore, (Xiao et al. 2025) examined generative adversarial network-based full-waveform inversion methodologies for quantitatively reconstructing hidden defects in high-density polyethylene pipe materials. Earlier, (Dasgupta et al. 2024) considered a novel modular inference approach combining two different generative models to approximate the posterior distribution of physics-based Bayesian inverse problems framed in high-dimensional ambient spaces. In other engineering applications, (Ye et al. 2024) established an auxiliary classification time series generation adversarial network to reflect the fluctuation characteristics of day-ahead wind power and power output level, while (Liu et al. 2024) examined a novel graph generative adversarial network for the accurate prediction of short-term future scenarios of a wind field. (Li et al. 2024) presented a genetic algorithm-based generative network to improve the design for urban wind conditions, (Zhang et al. 2024) provided a generative adversarial network for stochastic wind power output scenario generation, (Behara and Saha 2024) proposed a network to overcome reliability issues on wind and power conditions analysis, and (Li et al. 2024b) developed a transfer learning-based generative adversarial network for reconstructing inner-core high winds from synthetic aperture radar images. Importantly, an increasing interest is shown for using generative adversarial networks for structural optimization tasks (Ramu et al. 2022; Zhang et al. 2022; Yu et al. 2019; Tan et al. 2020; Qian and Ye 2021; Yonekura et al. 2022; Ates and Gorguluarslan 2021; Kim et al. 2022).

**Table 1** Z24 Bridge damage states (Garibaldi et al. 2003)

| Test | Description | Test | Description |
|---|---|---|---|
| PDT1 | 1st ref. measurement | PDT9 | Chipping of concrete, 12 m$^2$ |
| PDT2 | 2nd ref. measurement | PDT10 | Chipping of concrete, 24 m$^2$ |
| PDT3 | Settlement of pier, 20 mm | PDT11 | Landslide |
| PDT4 | Settlement of pier, 40 mm | PDT12 | Concrete hinges |
| PDT5 | Settlement of pier, 80 mm | PDT13 | Failure of anchor heads |
| PDT6 | Settlement of pier, 95 mm | PDT14 | Anchor heads #2 |
| PDT7 | Tilt of foundation | PDT15 | Rupture of tendons #1 |
| PDT8 | 3rd ref. measurement | PDT16 | Rupture of tendons #2 |
| | | PDT17 | Rupture of tendons #3 |

However, current damage detection-oriented methods based on generative adversarial networks focus only on data augmentation with known damage state or a known health state, which results in a supervised or a semi-unsupervised labeled process (Dunphy et al. 2022; Tilon et al. 2020; Li et al. 2024). For instance, (Soleimani-Babakamali et al. 2023) introduced the use of generative adversarial networks for dynamics learning after following a no-damage and damage data separate procedure, while (Luleci et al. 2023a) developed a supervised cyclic generative model to investigate the domain translation between undamaged and damaged acceleration data from one element to the same element as well as to other elements. (Yan et al. 2020) employed generative modeling only to re-balance the training dataset for chiller automatic fault detection and diagnosis, (Zhong et al. 2023) presented an improved deeper Wasserstein generative adversarial network with gradient penalty to generate datasets of pavement images, and (Guo et al. 2022) investigated a cyclic generative model to generate and discriminate surface damage images of conveyor belts. Furthermore, (Prajapati et al. 2025) developed a semi-supervised generative adversarial network model that can be trained on fewer samples based on a collection of Lamb wave interactions with a fiber reinforced composite plate under pristine and damaged conditions. (Luo et al. 2023) developed an unsupervised damage detection method that leverages improved generative adversarial network and cloud modeling which only needs the data in the healthy state of the structure for model training, without though justifying no-damage-related anomalies. Along these lines, (Lei et al. 2021) employed generative modeling for lost data reconstruction, (Mao et al. 2021) combined generative adversarial networks with autoencoders for anomaly detection, and the research is still ongoing (Shim et al. 2022; Rastin et al. 2021; Luleci et al. 2023).

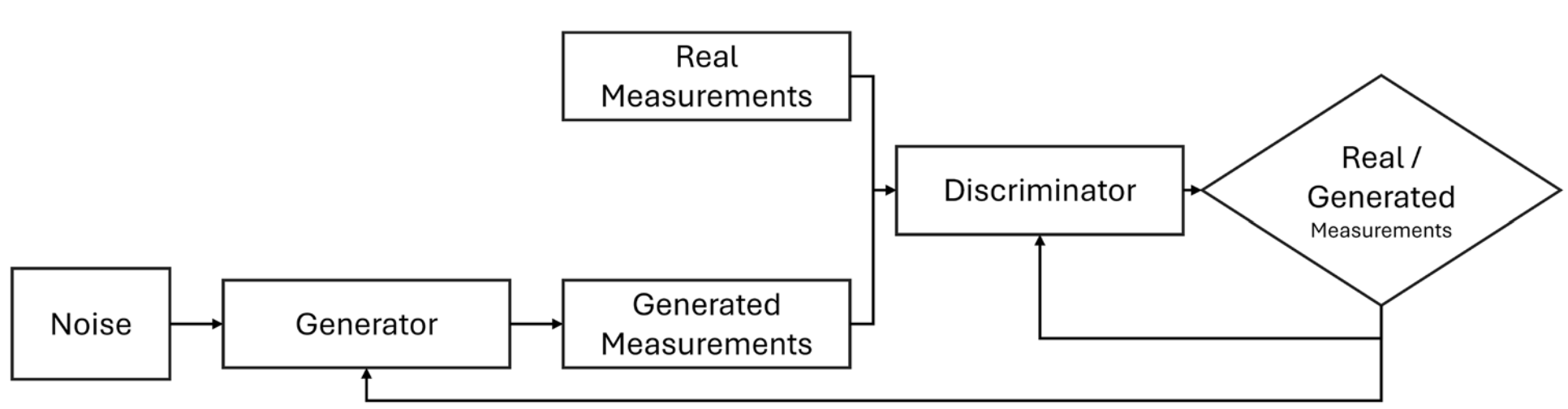

**Fig. 1** Generative adversarial network structure for structural damage detection and digital twinning

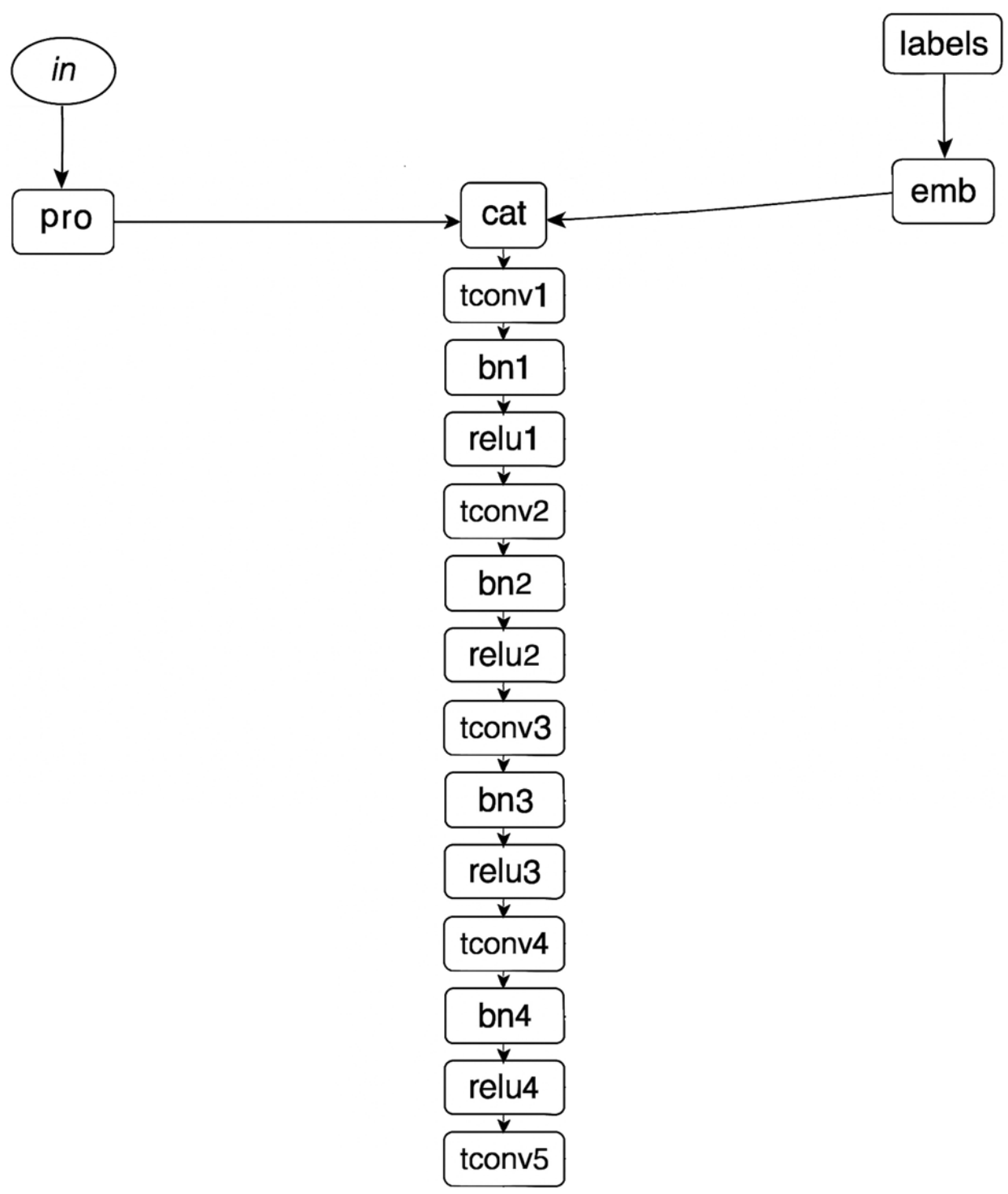

**Fig. 2** Examined generator network architecture for structural damage detection and digital twinning

**Table 2** Examined generator network architecture for structural damage detection and digital twinning

| Name | Type | Activations | Learnables |
|---|---|---|---|
| in | Image Input | 1(S) × 1(S) × 100(C) × 1(B) | 0 |
| proj | Project and Reshape | 4(S) × 1(S) × 1024(C) × 1(B) | 413,696 |
| labels | Image Input | 1(S) × 1(S) × 1(C) × 1(B) | 0 |
| emb | Reshape layer | 4(S) × 1(S) × 1(C) × 1(B) | 604 |
| cat | Concatenation | 1(S) × 1(S) × 1025(C) × 1(B) | 0 |
| tconv1 | 2-D Transposed Convolution | 8(S) × 1(S) × 512(C) × 1(B) | 2,624,512 |
| bn1 | Batch Normalization | 8(S) × 1(S) × 512(C) × 1(B) | 1024 |
| relu1 | ReLU | 8(S) × 1(S) × 512(C) × 1(B) | 0 |
| tconv2 | 2-D Transposed Convolution | 36(S) × 1(S) × 256(C) × 1(B) | 1,310,976 |
| bn2 | Batch Normalization | 36(S) × 1(S) × 256(C) × 1(B) | 512 |
| relu2 | ReLU | 36(S) × 1(S) × 256(C) × 1(B) | 0 |
| tconv3 | 2-D Transposed Convolution | 150(S) × 1(S) × 128(C) × 1(B) | 393,344 |
| bn3 | Batch Normalization | 150(S) × 1(S) × 128(C) × 1(B) | 256 |
| relu3 | ReLU | 150(S) × 1(S) × 128(C) × 1(B) | 0 |
| tconv4 | 2-D Transposed Convolution | 599(S) × 1(S) × 64(C) × 1(B) | 41,024 |
| bn4 | Batch Normalization | 599(S) × 1(S) × 64(C) × 1(B) | 128 |
| relu4 | ReLU | 599(S) × 1(S) × 64(C) × 1(B) | 0 |
| tconv5 | 2-D Transposed Convolution | 1201(S) × 1(S) × 1(C) × 1(B) | 449 |

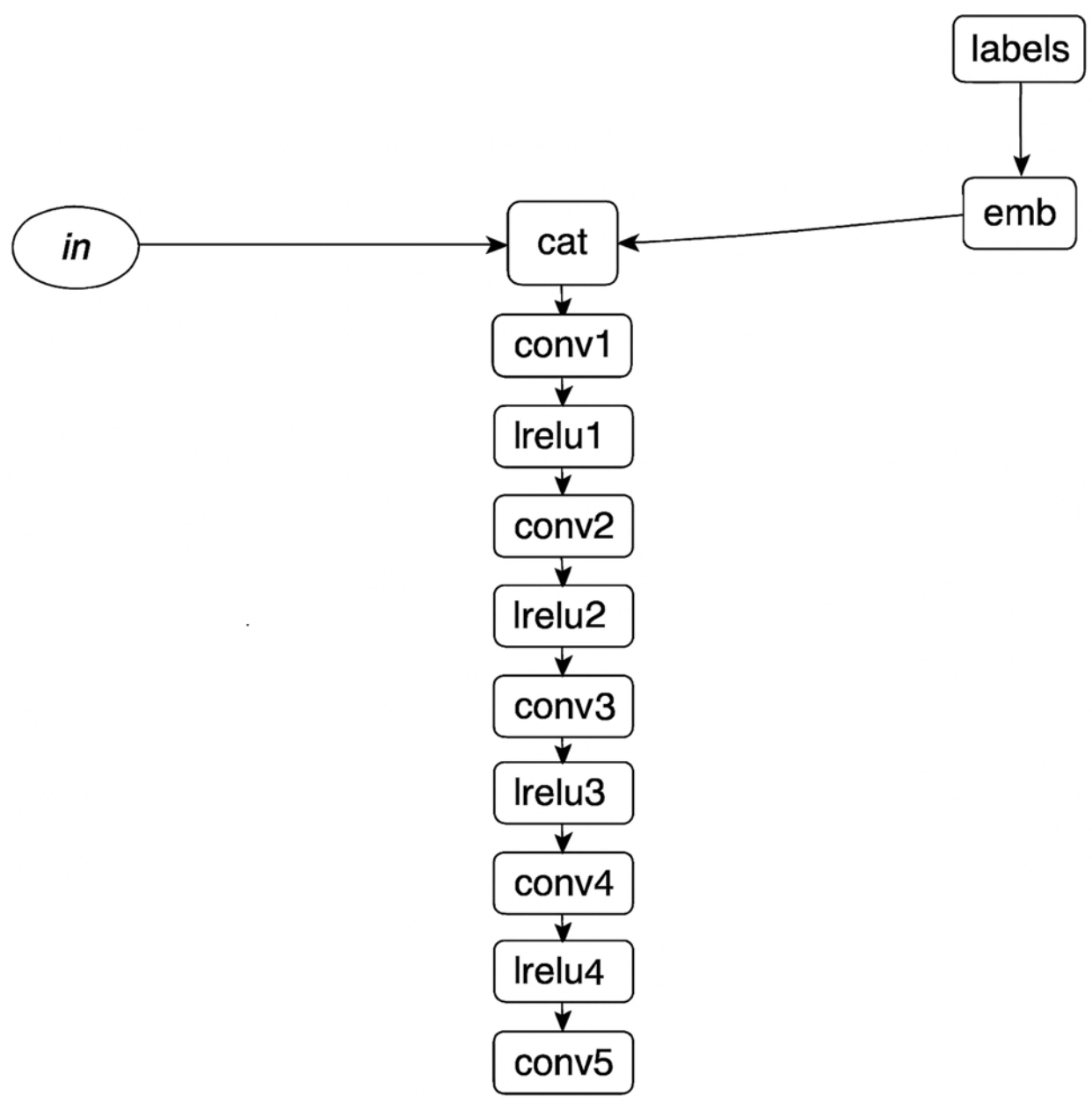

**Fig. 3** Examined discriminator network architecture for structural damage detection and digital twinning

**Table 3** Examined discriminator network architecture for structural damage detection and digital twinning

| Name | Type | Activations | Learnables |
|---|---|---|---|
| in | Image input | 1201(S) × 1(S) × 1(C) × 1(B) | 0 |
| labels | Image input | 1(S) × 1(S) × 1(C) × 1(B) | 0 |
| emb | Reshape Layer | 1201(S) × 1(S) × 1(C) × 1(B) | 121,501 |
| cat | Concatenation | 1201(S) × 1(S) × 2(C) × 1(B) | 0 |
| conv1 | 2-D Convolution | 594(S) × 1(S) × 512(C) × 1(B) | 17,920 |
| lrelu1 | Leaky ReLU | 594(S) × 1(S) × 512(C) × 1(B) | 0 |
| conv2 | 2-D Convolution | 146(S) × 1(S) × 256(C) × 1(B) | 2,097,408 |
| lrelu2 | Leaky ReLU | 146(S) × 1(S) × 256(C) × 1(B) | 0 |
| conv3 | 2-D Convolution | 34(S) × 1(S) × 128(C) × 1(B) | 524,416 |
| lrelu3 | Leaky ReLU | 34(S) × 1(S) × 128(C) × 1(B) | 0 |
| conv4 | 2-D Convolution | 8(S) × 1(S) × 64(C) × 1(B) | 65,600 |
| lrelu4 | Leaky ReLU | 8(S) × 1(S) × 64(C) × 1(B) | 0 |
| conv5 | 2-D Convolution | 1(S) × 1(S) × 1(C) × 1(B) | 513 |

To this end, this work introduces a novel concept within conditional generative adversarial networks approach (Mirza and Osindero 2014) for simultaneous structural damage detection and digital twinning, without prior knowledge of damage existence. The approach uses the training period where the networks converge to quantify the additional dynamics learnt as an indicator. This indicator implies changes in the system dynamics. Specifically, the approach explores the convergence dynamics of the model during training as a novel, unsupervised indicator of structural novelty. The core hypothesis is that greater structural changes (e.g., from healthy to severely damaged) require longer training convergence, revealing distribution mismatches and hidden nonlinearities. This insight allows for exploiting the training behavior itself as a damage-sensitive metric. Firstly, unknown damage state measurements are used as an input, while the model is forced to converge conditionally to two different damage states. Secondly, the process is repeated for a different group of measurements. Finally, the convergence scores are compared to identify which group belongs to different damage states. The process creates simultaneously measurements for digital twinning purposes at different damage states, capable of pattern recognition and machine learning data generation. Further to this process, a support vector machine classifier and a principal component analysis procedure is developed to assess generated and real measurements of each damage category as a secondary indicator of satisfactory dynamics learning. The proposed method results in vibration response generation conditioned on specific structural damage states, healthy or damaged, enabling the transformation and interpolation between different structural configurations. It is applied to the benchmark Z24 Bridge dataset, a full-scale monitoring experiment comprising progressive damage scenarios over a year-long period. Unlike existing generative adversarial network-based damage detection methods that rely on labeled datasets or known damage scenarios, the proposed approach employs a conditional-labeled structure to enable unsupervised learning. This allows the model to not only detect damage without prior labeling but also generate and recognize novel damage states.

The work is organized as follows: Section 2 provides the concept of conditional generative adversarial networks for damage detection using their training learning duration. Section 3 presents the neural network model used in this work, including all parameters and architectures. Section 4 applies the method to the Z24 Bridge for both healthy-to-healthy measurements and healthy-to-damaged measurements. It also includes the classification results using the support vector machines algorithm, as well as generated measurements and spectra. The discussion, limitations, and future work are presented in Section 5. Finally, Section 6 concludes the work.

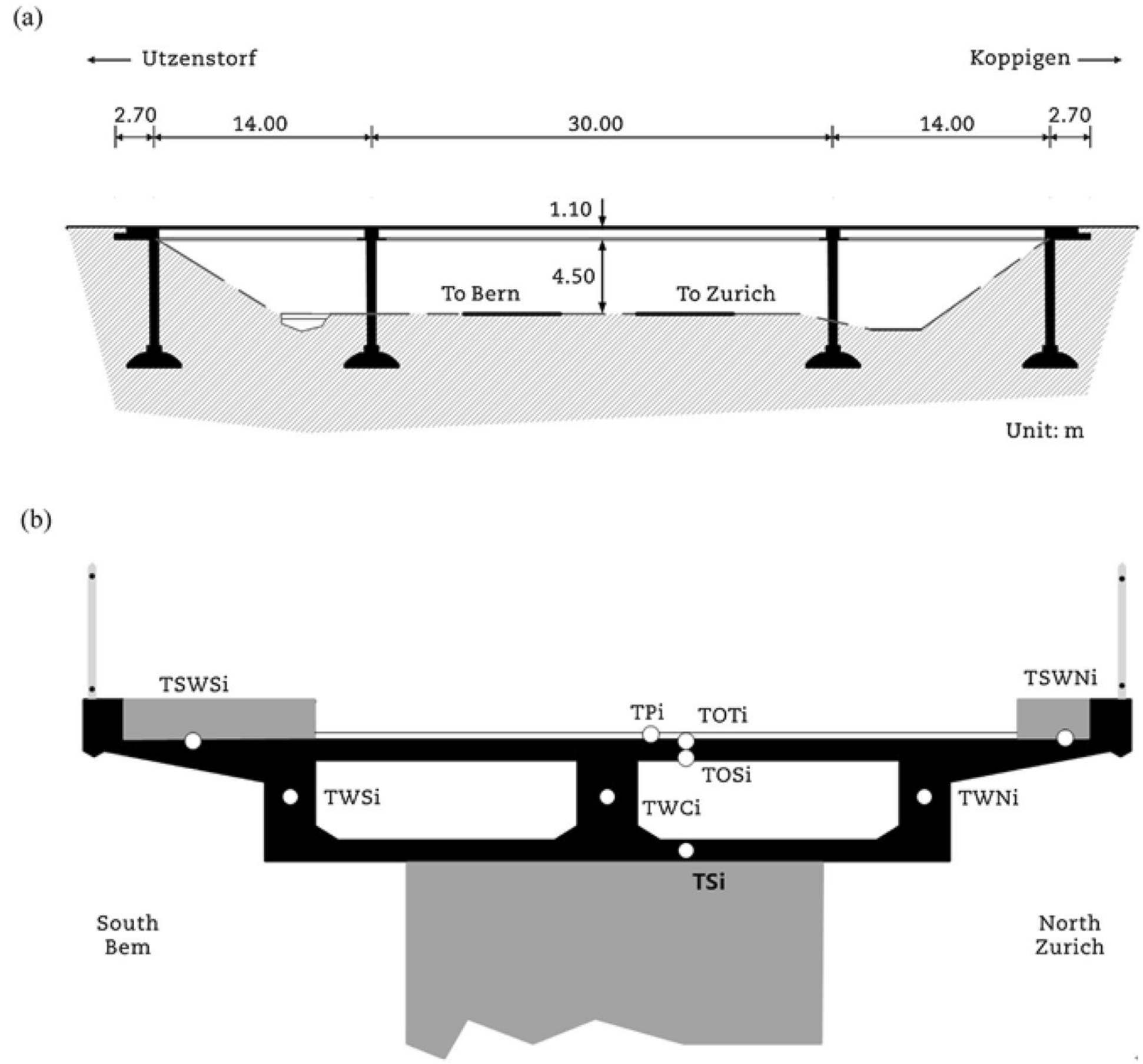

**Fig. 4** Horizontal view (a) and cross-sectional view (b) of the Z24 Bridge and location of the thermocouples by the variable i for the span number (Peeters and De Roeck 2001) (research permission by KU Leuven)

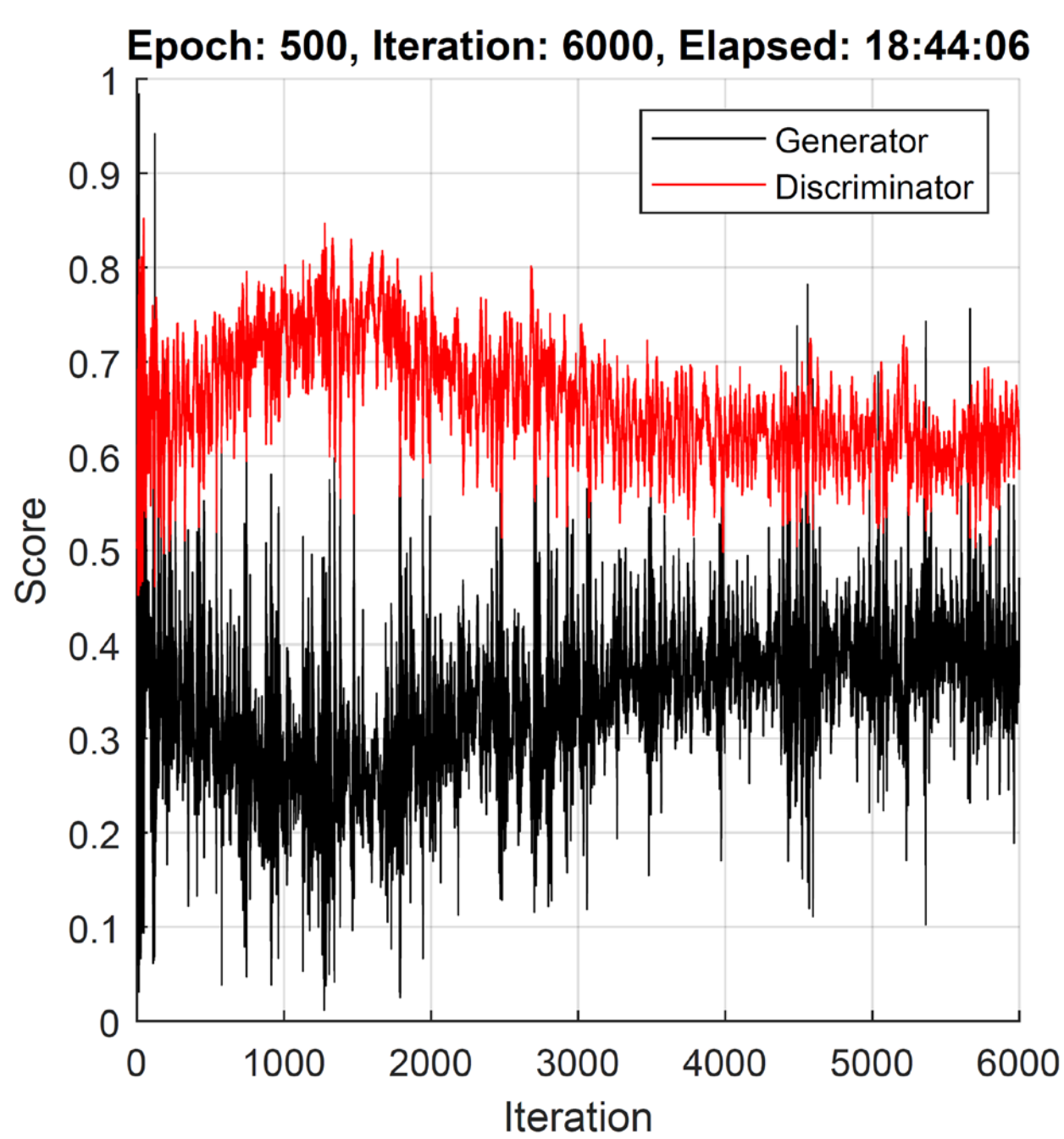

**Fig. 5** Score for both networks when trained using healthy PDT1 to damage PDT8 measurements for $500^{th}$ epochs

## 2 Damage detection and digital twinning using generative adversarial networks

Generative adversarial networks are a class of deep learning models designed for generating data which closely resembles real measurements. Introduced by (Goodfellow et al. 2014), they consist of two neural networks, the generator and the discriminator, which compete in a minimax procedure; see Fig. 1.

Specifically, on the one side, the generator maps a random noise vector from a latent space to the data domain, producing generated measurements. On the other side, the discriminator tries to differentiate between real measurements and generated ones. Finally, the networks are trained iteratively, where the generator aims to improve the quality of generated measurements, while the discriminator refines its ability to distinguish real from generated measurements. The objective function of a standard model is formulated as:

$$\min_G \max_D V(D, G) = \mathbb{E}_{x \sim p_{\text{data}}(x)}[\log D(x)] + \mathbb{E}_{z \sim p_z(z)}[\log(1 - D(G(z)))] \tag{1}$$

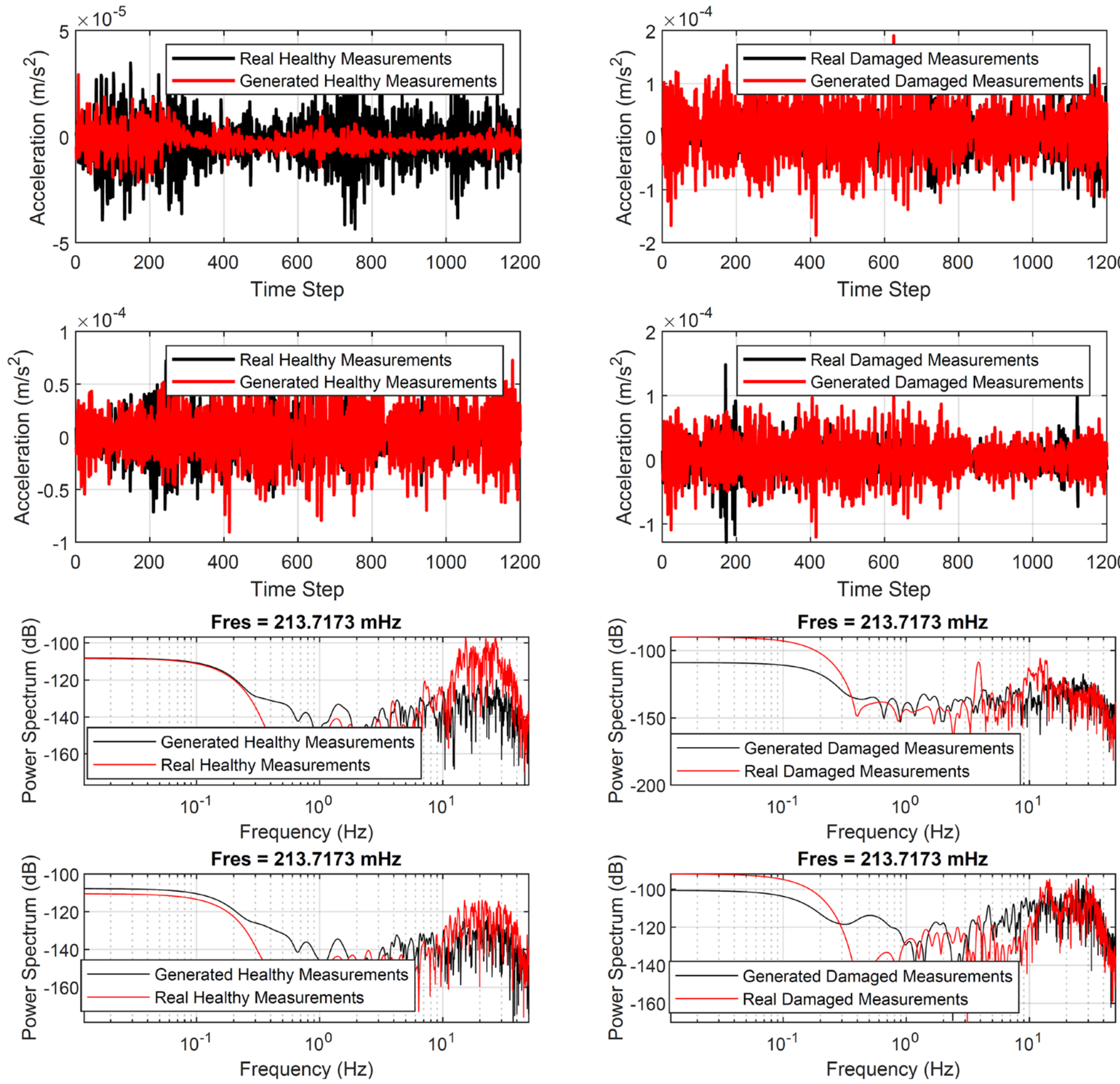

**Fig. 6** Time- and frequency-domain generated measurements for healthy PDT1 to damage PDT8 measurements for $500^{th}$ epochs

where $x \sim p_{\text{data}}(x)$ represents real measurements, $z \sim p_z(z)$ is a random noise input to the generator, $G(z)$ are the generated measurements, $D(x)$ is the probability that $x$ is real, $D(G(z))$ is the probability that the generated measurements are real. Here, the discriminator $D$ is trained to maximize $\log D(x)$ to correctly classify real measurements, and $\log(1 - D(G(z)))$ to correctly classify generated measurements. The generator $G$ is trained to minimize $\log(1 - D(G(z)))$ to generate measurements that maximize $D(G(z))$, effectively convincing the discriminator. Often, to avoid vanishing gradients, the generator is often trained using the loss:

$$\max_G \mathbb{E}_{z\sim p_z(z)}[\log D(G(z))] \tag{2}$$

which is equivalent to minimizing:

$$\mathbb{E}_{z\sim p_z(z)}[\log(1 - D(G(z)))] \tag{3}$$

which provides better gradient behavior during training.

To generate healthy and damage structural monitoring measurements, a conditional generative adversarial network is employed (Mirza and Osindero 2014). The objective function is now modified as follows:

$$\begin{aligned}\min_G \max_D V(D, G) = \mathbb{E}_{x\sim p_{\text{data}}(x|y)}[\log D(x|y)] \\ + \mathbb{E}_{z\sim p_z(z)}[\log(1 - D(G(z|y)|y))]\end{aligned} \tag{4}$$

where now, $x \sim p_{\text{data}}(x|y)$ represents real measurements conditioned on auxiliary information $y$ whether the structure is damaged or not, $z \sim p_z(z)$ is a random noise input to the generator, $G(z|y)$ are the generated measurements conditioned on $y$, $D(x|y)$ is the probability that $x$ is real given $y$, $D(G(z|y)|y)$ is the probability that the generated measurements are real given $y$. Here, the training dataset comprises measurements from both healthy and damaged structure, where the generator learns to generate measurements that maintain statistical and frequency properties similar to the real healthy and damaged measurements, while the discriminator ensures that generated measurements are indistinguishable from actual experimental recordings. Initially, all measurements are preprocessed and normalized, before the generator employs a deep neural network with fully connected and convolutional layers to model temporal dependencies. The networks are then trained iteratively using the Adam optimizer (Kingma 2014; Impraimakis 2025).

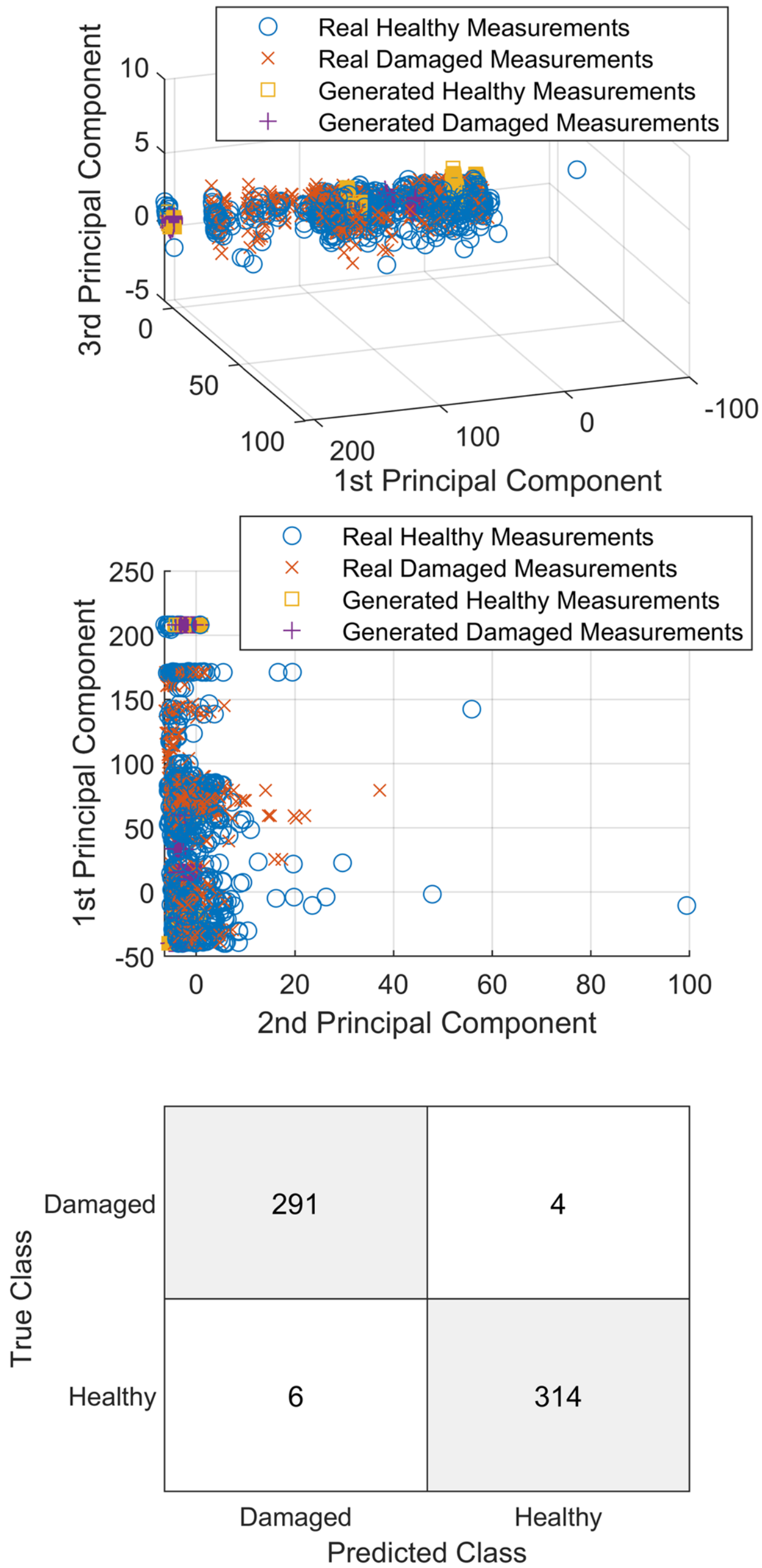

**Fig. 7** Real and generated measurements healthy PDT1 to damage PDT8 measurements on the first three principal components and support vector machine classifier performance for $500^{th}$ epochs

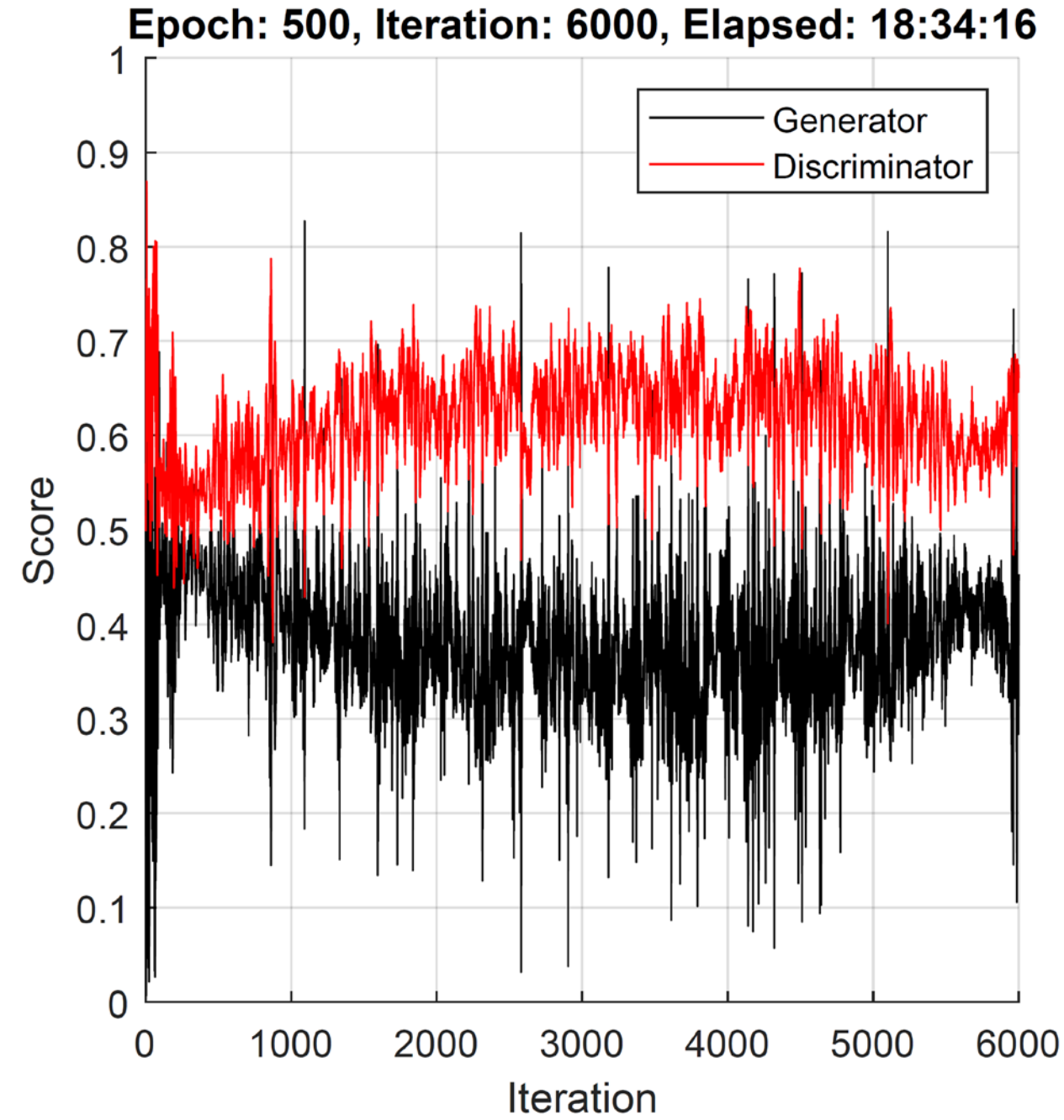

**Fig. 8** Score for both networks when trained using healthy PDT1 to damage PDT17 measurements for $500^{th}$ epochs

Finally, the generated measurements are compared against real measurements using statistical similarity measures.

The conditional model is then applied to an end-to-end global structure health monitoring system to map signals from one damage state to another to investigate how well it can learn the transformation between different structural conditions. The scenario where the model converges faster correlates with structural similarity; namely, the source and target states are similar, and there is no significant damage evolution. Alternatively, the lengthier convergence learning process provides an indication of progressive or novel damage. Specifically, the convergence behavior of the conditional generative adversarial network is mathematically quantified by measuring the length or area under the training curve above a specified threshold, which reflects the dynamic learning period. This area varies systematically with different structural damage states, offering a mathematical indicator of the model's adaptation to varying signal distributions. The final goal is to exploit this behavior to detect and quantify structural degradation using unsupervised learning.

## 3 Examined conditional generative adversarial network model

To apply the method to a real-world problem, the Z24 Bridge dataset is used as a well-established benchmark for structural health monitoring, containing vibration response data from a real bridge subjected to progressive damage over time (Teughels and De Roeck 2004; Peeters and De Roeck 2001; Maeck and De Roeck 2003; Garibaldi et al. 2003; Maeck and De Roeck 2003a). The Z24 Bridge was a post-tensioned concrete highway bridge located near Koppigen, Switzerland. It was part of a full-scale monitoring and controlled damage experiment conducted by the Swiss

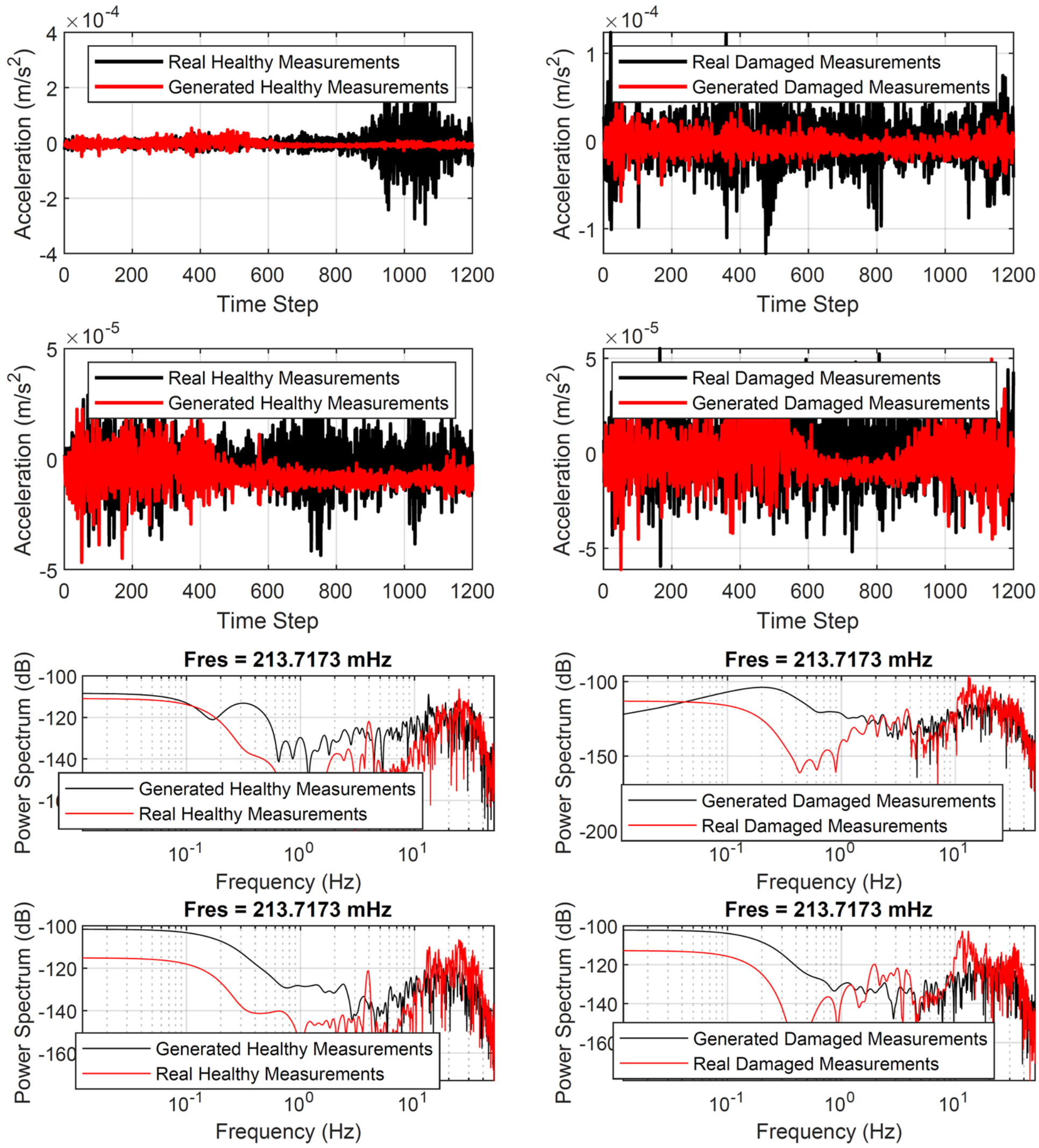

**Fig. 9** Time- and frequency-domain generated measurements for healthy PDT1 to damage PDT17 measurements for 500$^{th}$ epochs

Federal Laboratories for Materials Science and Technology before the bridge's planned demolition. The dataset collection spans approximately one year and includes acceleration response measurements under ambient excitation. During this period, 17 damage scenarios were introduced systematically, such as hinge degradation and tendon cutting as given in Table 1, where PDT stands for progressive damage state. More details about the utilized data is provided in Section 5, Table 4.

The examined generator is designed to map a random noise vector and conditional information to a realistic output. It consists of multiple fully connected and convolutional layers to capture temporal dependencies. The architecture is shown in Fig. 2 and in Table 2, proving also the spatial dimension/size (S), the batch size (B), and the channels/features (C). The discriminator is also a convolutional network that classifies input signals as real or generated. The architecture is shown in Fig. 3 and in Table 3. Both networks are trained using the Adam optimizer with a learning rate of 0.0005, number of epochs 500, and mini-batch size of 128. The loss function follows the standard binary cross-entropy formulation for adversarial training.

## 4 Application to Z24 Bridge structural health monitoring measurements

A schematic representation of the dynamic system is shown in Fig. 4 by (Peeters and De Roeck 2001). Figure 5 refers to the case where measurements of healthy-to-damaged state are utilized to train the model. It shows the score of both networks converging roughly to 0.5, meaning that an equilibrium has been achieved when no network is more powerful than the other. This application, specifically, examines the learning duration when measurements from PDT01

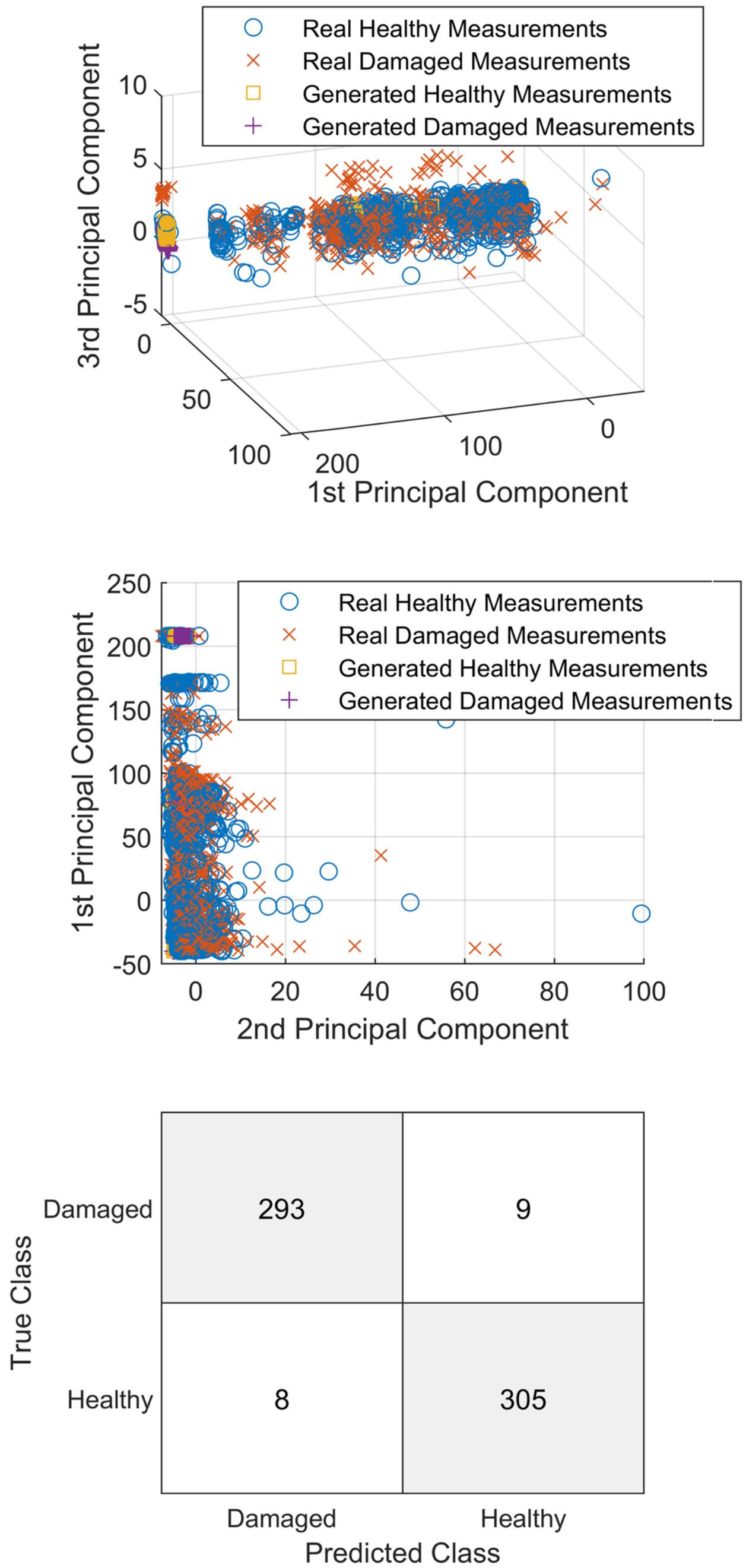

Fig. 10 Real and generated measurements healthy PDT1 to damage PDT17 measurements on the first three principal components and support vector machine classifier performance for $500^{th}$ epochs

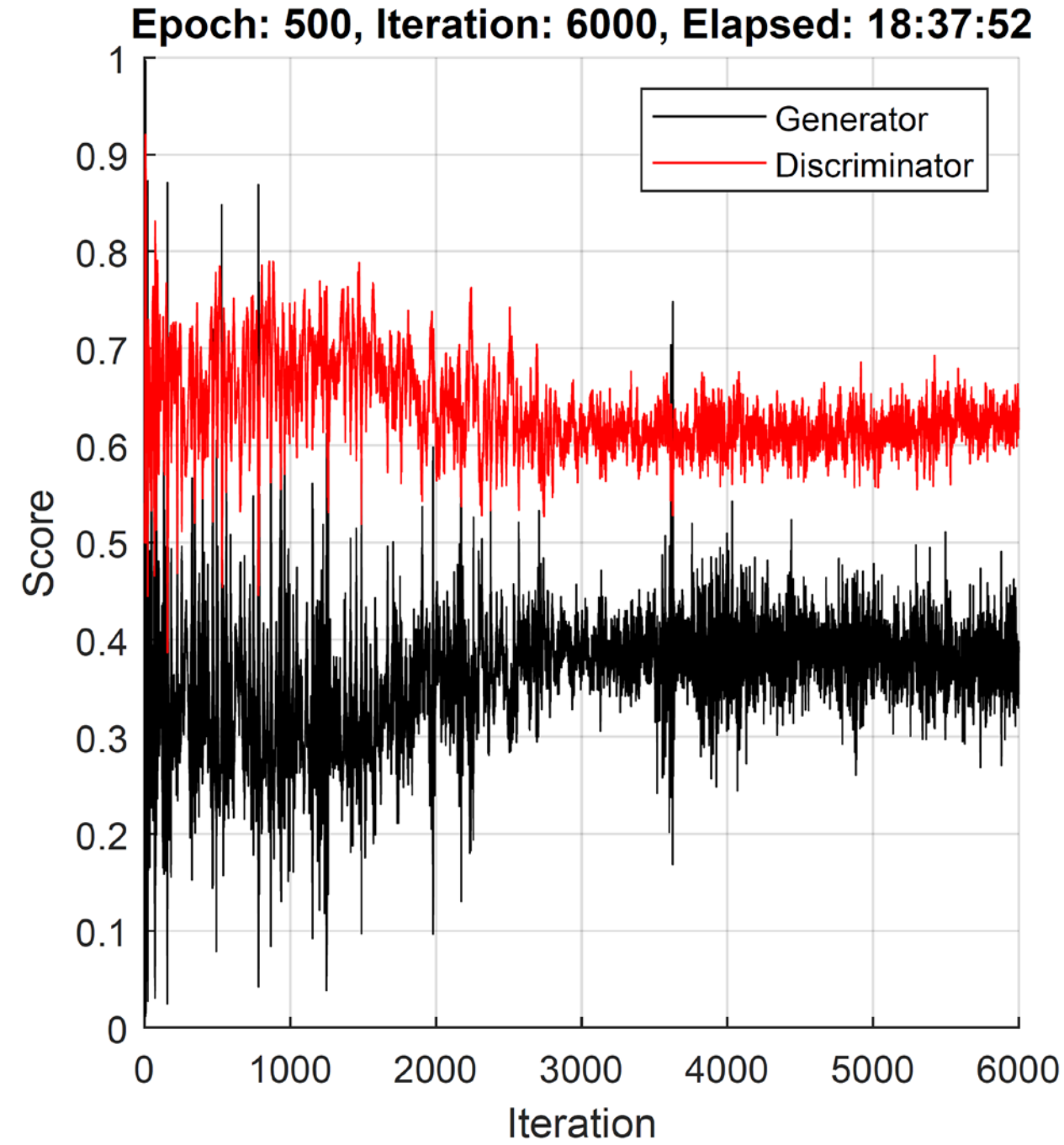

Fig. 11 Score for both networks when trained using healthy PDT1 to healthy PDT1 (faked "Damaged") of a different configuration measurements for $500^{th}$ epochs

and PDT08 are considered. This behavior is shown for the last $500^{th}$ epoch, after 6000 iterations of total duration 18 h, 44 min, and 06 s of training on a computer with an Intel Core Ultra 7 155U processor and 16GB of RAM. The training time can be significantly reduced when using graphics processing units. The score shows that for a large number of epochs the models are still learning, indicating that there is a meaningful distinction between the measurements (roughly 3000 iterations of intense learning).

After the model has been trained, it is now available to provide new generated data shown in Fig. 6 of both healthy and damaged measurements. Here, 615 new measurement signals are generated. Unlike images and audio signals, vibration signals have characteristics that make them difficult for human perception to directly distinguish them. To compare real and generated measurements, the principal component analysis (PCA) is applied to derive features (mean value, variance, dominant frequencies, autocorrelation, etc.) of the real measurements, and then project the same features of the generated measurements to the same PCA subspace.

Figure 7, then, shows the real and the generated measurements samples of the same category which lie in the same areas of the first three principal components. Namely, accurate and inaccurate measurements lie in the same area of the PCA subspace regardless of their being real or generated, demonstrating that the generated measurements have features similar to those of the real measurements. Distinction between healthy and damaged data is not clear though in the PCA plotting.

To further illustrate the performance of the model, a support vector machine classifier is utilized based on the generated measurements to predict whether a real measurement is accurate or inaccurate. Initially, the generated measurements

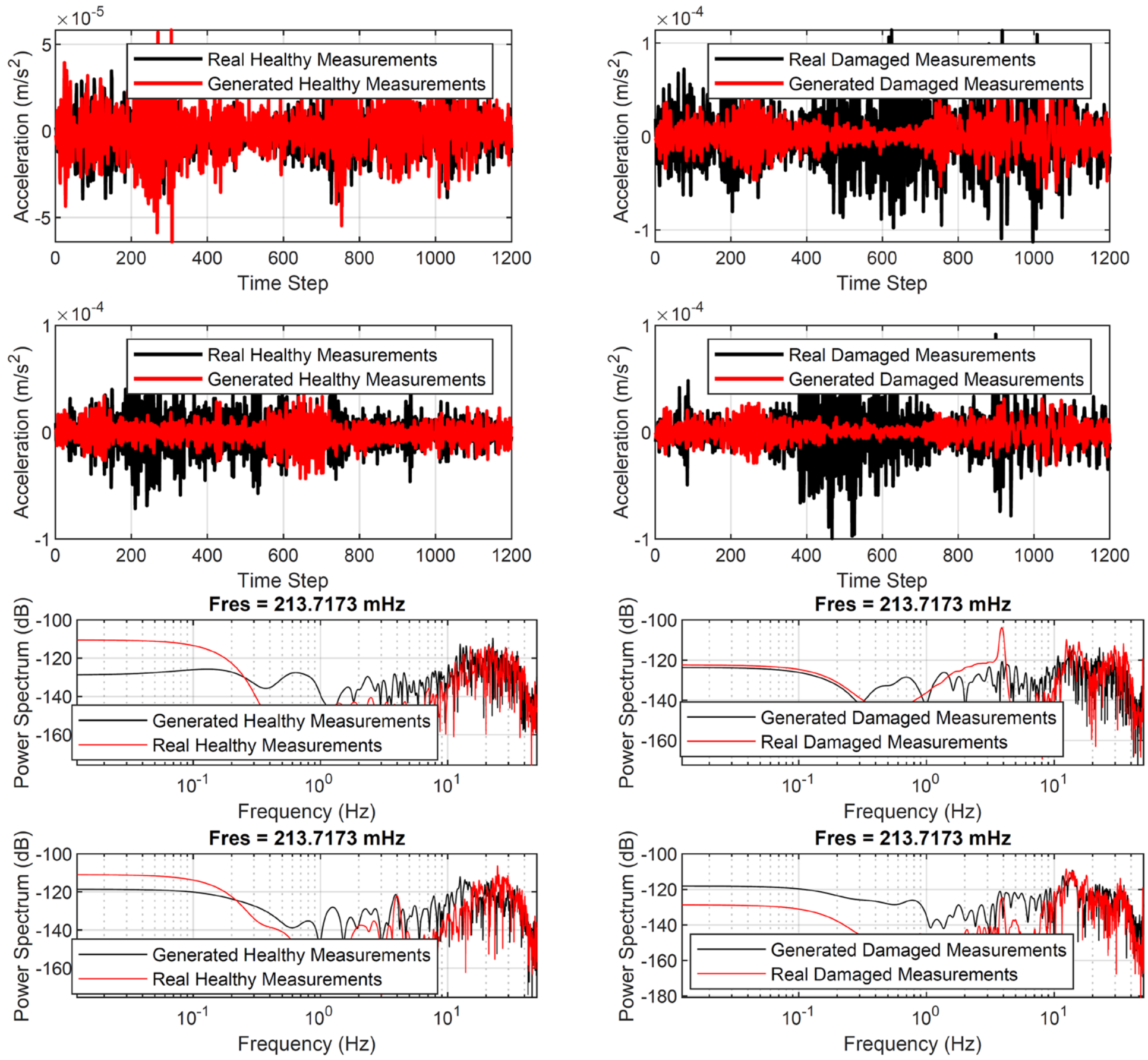

**Fig. 12** Time- and frequency-domain generated measurements for healthy PDT1 to healthy PDT1 (faked "Damaged") of a different configuration measurements for $500^{th}$ epochs

are set as the training dataset, while the real measurements as the test dataset. After the classifier has been trained, it is used to obtain the predicted category for the real measurements. In this application, the classifier achieves a prediction accuracy above 90%. Finally, in Fig. 7 (bottom plot) a confusion matrix is provided for the prediction performance for each category, where the classifier trained on the generated measurements achieves a high degree of accuracy.

Along these lines, Fig. 8 refers to the same process but when PDT1 is compared to PDT17. Here, the model shows an even lengthier learning duration. This indicates a richer dynamics learning scenario due to a heavier damage for the structure. The score shows that for a large number of epochs, the models are still learning which indicates that there is a meaningful distinction between the measurements (roughly 5000 iterations of intense learning). Figures 9 and 10 show, in a similar manner to Fig. 6 and 7, generated measurements and classification accuracy, which exceeds 90%.

On the other hand, an alternative investigation is provided when the measurements are compared for the same damage state, but in a different day or in different sensor configuration. Figure 11 refers to the same process but when PDT1 is compared to a different PDT1 configuration, namely 01setup01 vs 01setup02 (Peeters and De Roeck 2001). Here, the model shows a much lower learning period compared to Figs. 5 and 8. This indicates that not much new dynamics is learnt due to no further damage existence for the structure (roughly 2000 iterations of intense learning). Figures 12 and 13 show the generated measurements and the classification accuracy for this approach which unrealistically approaches 100%, indicating that the model is learning a fake unrealistic dynamics. As a result, either the model scoring process can give indications to whether damage actually exists, or the unrealistic classification performance.

Furthermore, Fig. 14 repeats such a scenario but for when the PDT8 test is compared to a different PDT8 configuration, namely 08setup01 vs 08setup02 (Peeters and De Roeck 2001). Once more, the model shown a much lower learning period, indicating that not much new dynamics is learnt due to no further damage state of the structure (no clear learning period). The classification accuracy, once more, is approximating 100% which indicates that the model is learning a fake unrealistic dynamics in the generated measurements in Fig. 15 and the classification accuracy in Fig. 16.

Finally, Fig. 17 repeats such the test but for when PDT17 is compared to a different PDT17 configuration with no clear learning period, with generated measures in Fig. 18 and classification accuracy in Fig. 19, with the same conclusion

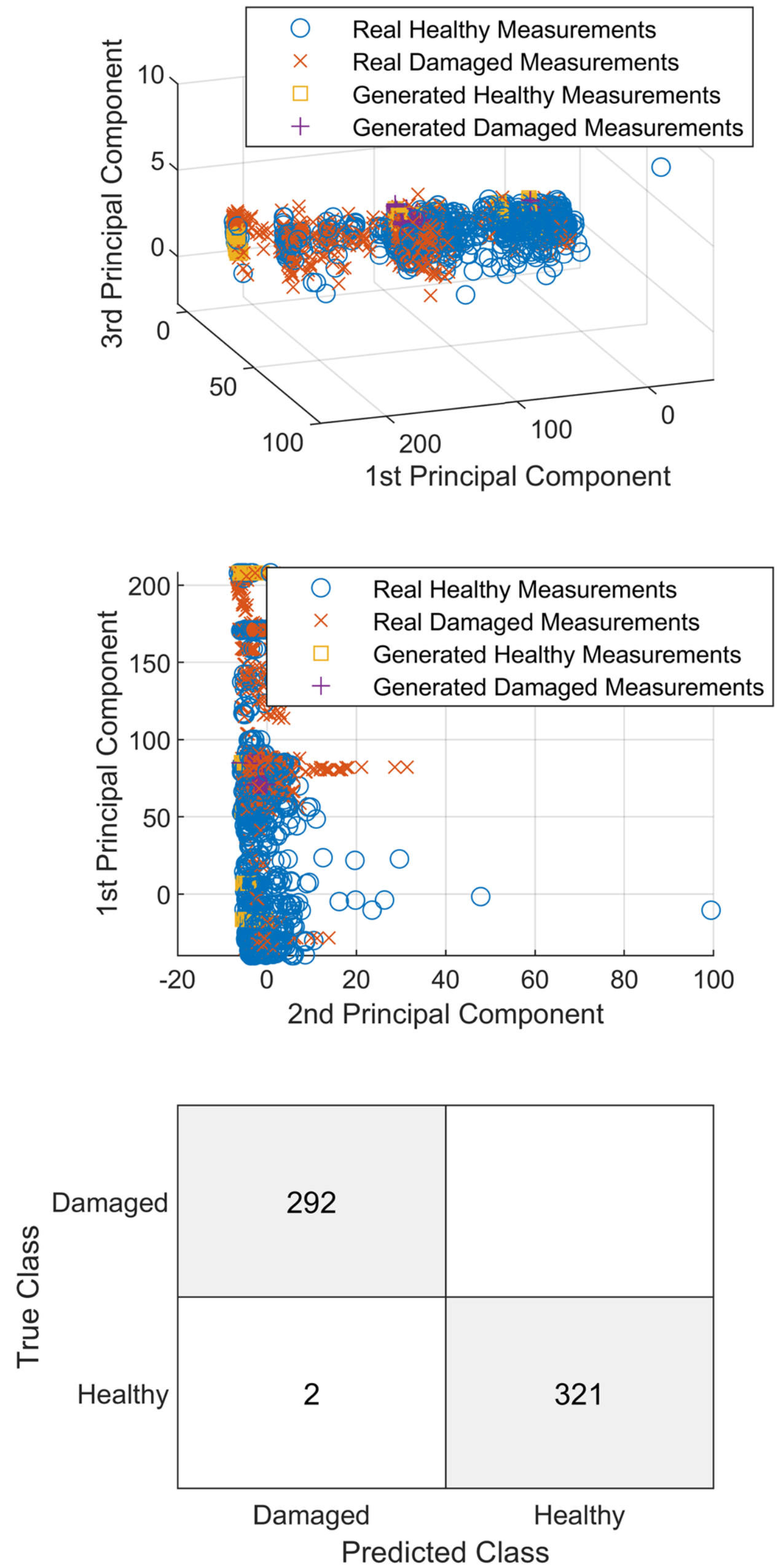

**Fig. 13** Real and generated measurements healthy PDT1 to healthy PDT1 (faked "Damaged") of a different configuration measurements on the first three principal components and support vector machine classifier performance for $500^{th}$ epochs

when using 17setup01 vs 17setup02 (Peeters and De Roeck 2001).

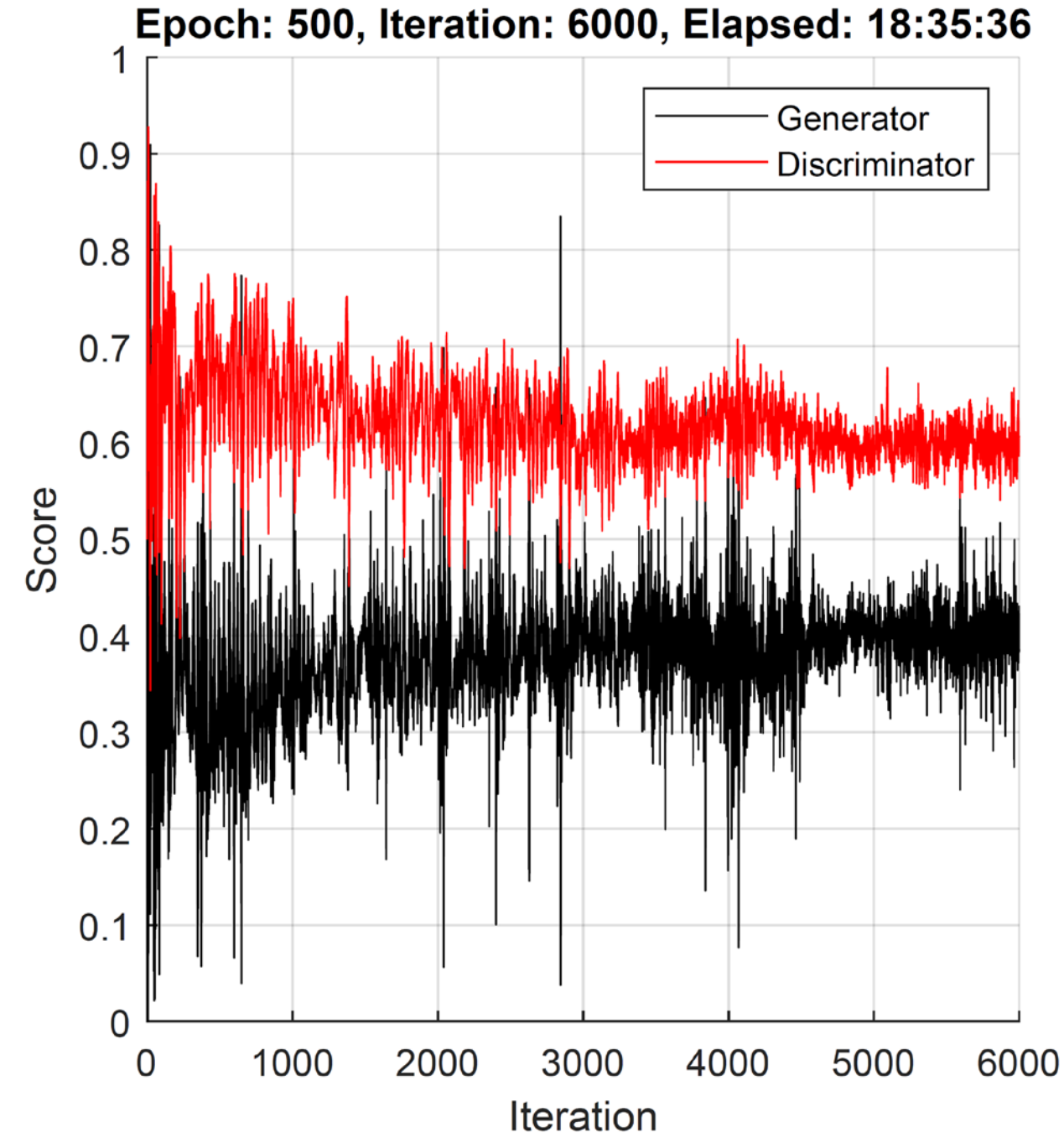

**Fig. 14** Score for both networks when trained using damaged PDT8 (faked "Healthy") to damaged PDT8 of a different configuration measurements for $500^{th}$ epochs

## 5 Discussion

This work provided a simple and effective way for simultaneous damage detection and structural response generation, without a prior knowledge of the damage state. In healthy-to-healthy or same damage-level measurements scenario, the networks scores show a faster convergence behavior. In the damaged-to-healthy scenario, though, more time was needed for convergence as more dynamics was being learnt. This insight is an important benefit in damage detection processes, as prior knowledge often has a significant effect on models' predictions (Beck 2010; Huang et al. 2017; Impraimakis and Smyth 2022a; Jacobs et al. 2018), equal to the hyper-parameter or noise parameter tuning (Yuen et al. 2022; Gres et al. 2025; Teymouri et al. 2023; Bilgin and Olivier 2025; Kontoroupi and Smyth 2016).

A potential concern is related to using the generative model learning duration as an anomaly detector, which may not necessarily justify a full proof of damage. It is, though, a strong indicator that the signal has structural or dynamic changes, which may correspond to damage. This is especially true if the pattern of change is consistent across multiple samples or sensors, or when damage introduces nonlinearity, loosened joints, or cracks, which often allow for higher amplitude vibrations, introduction of new frequency

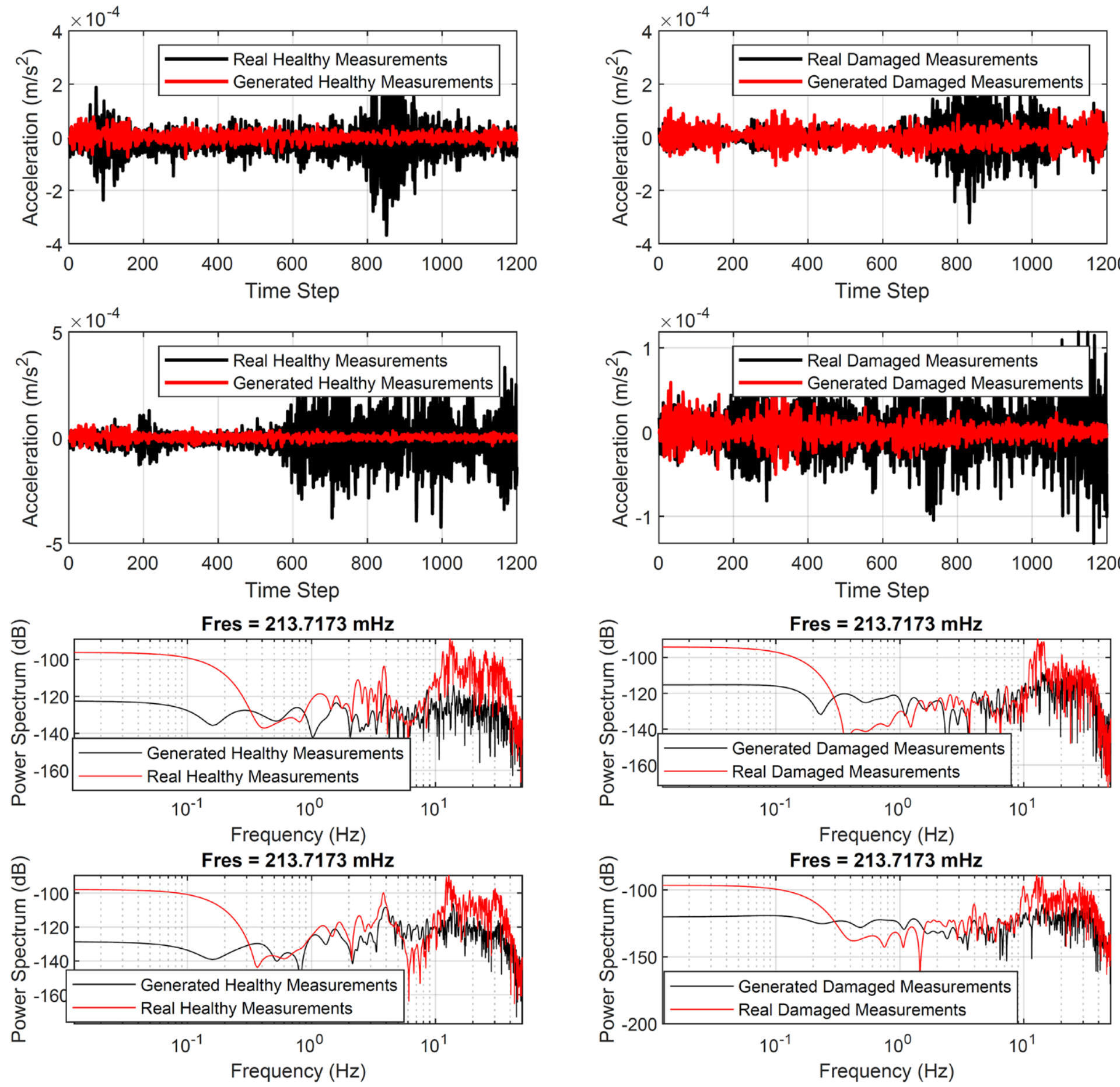

**Fig. 15** Time- and frequency-domain generated measurements for damaged PDT8 (faked "Healthy") to damaged PDT8 of a different configuration measurements for $500^{th}$ epochs

components, or reduction in damping (Impraimakis and Smyth 2022b; Impraimakis 2024). In this case though, the data should be normalized the same way (same units, no gain mismatch), so that the change is not caused by external factors such as different loadings, excitation conditions or sensor dysfunctions. One should analyze multiple damaged samples to see whether this is repeatable.

The fundamental basis behind the generative adversarial network approach is that the model is essentially trying to project damaged data back into healthy space, which is harder, hence slower convergence. Slower convergence implies distribution mismatch between damaged and healthy. Training difficulty becomes a surrogate for anomaly or novelty detection. A potential metric for this behavior is the number of iterations until the discriminator loss stabilizes, or the area under the generator/discriminator loss curves. It is important to note, though, that generative adversarial networks are notoriously unstable (Wiatrak et al. 2019; Sajeeda and Hossain 2022; Thanh-Tung et al. 2019); slow convergence could also mean bad initialization or mode collapse risk. Therefore comparison of the same model and architecture, same input, and with the same hyper-parameters should be followed.

Related to the data used in this study, the measurements originated from the Z24 Bridge monitoring dataset, specifically from ambient vibration tests conducted under three progression steps: PDT01 (01setup01), PDT08 (08setup01), and PDT17 (17setup01) as shown in Fig. 20. All data included 33 sensors recorded at a sampling rate of 100 Hz, yielding a total of 65,536 time steps per measurement. To facilitate efficient model training and evaluation, the long raw signals were subdivided into 787 shorter signals of 1201 time steps each. No overlapping window was applied during this segmentation. All signals were preprocessed using a detrending operation to remove linear components, followed by normalization in which the mean was subtracted and the result divided by the standard deviation. A generative model was then trained to synthesize data, and 1000 signals were generated for each class, using the same sampling ratio. The detailed summary of the dataset preparation and segmentation is provided in Table 4.

To demonstrate the feasibility of the method when using multiple damage state scenarios instead of a binary

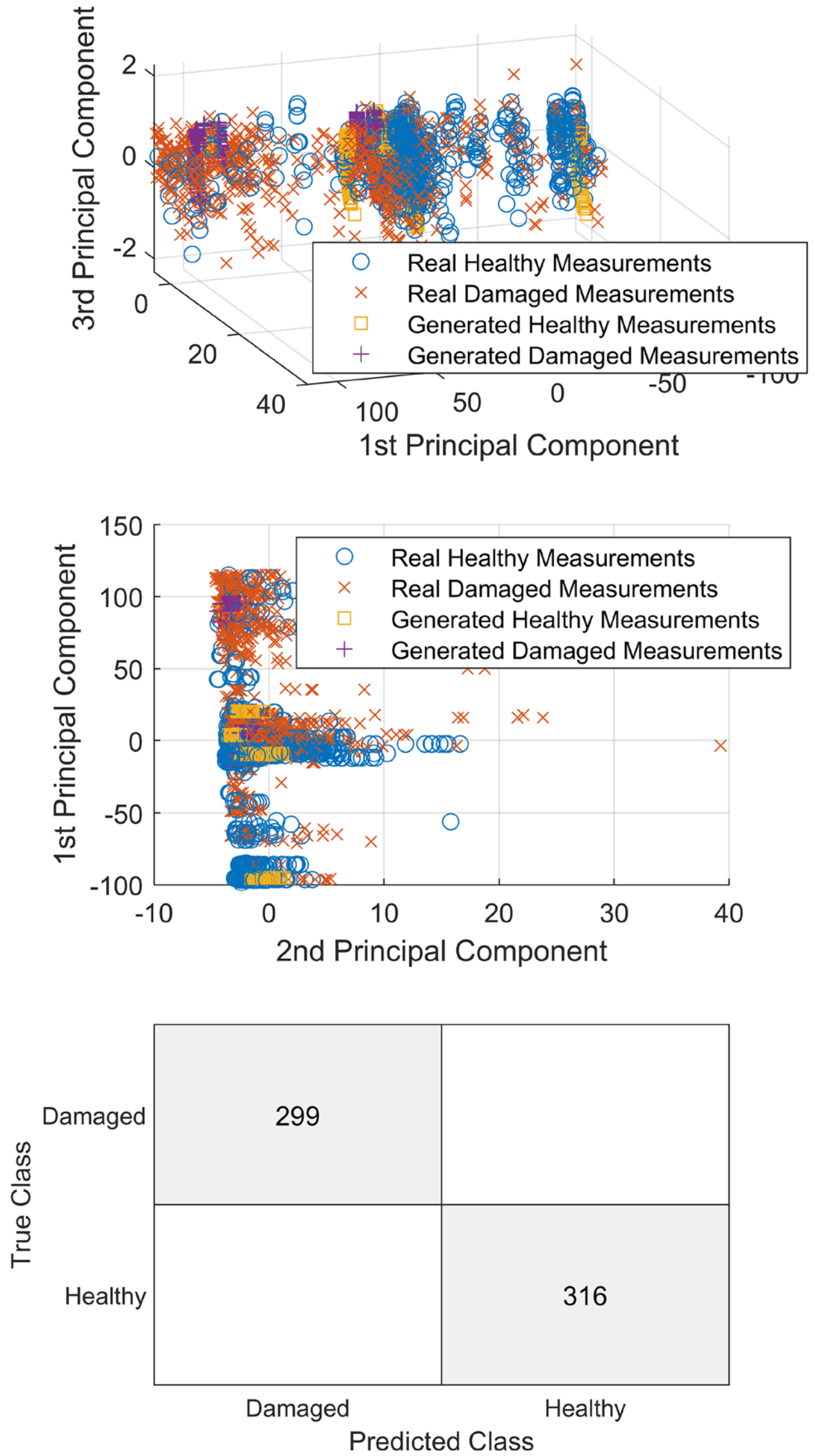

Fig. 16 Real and generated measurements damaged PDT8 (faked "Healthy") to damaged PDT8 of a different configuration measurements on the first three principal components and support vector machine classifier performance for 500$^{th}$ epochs

classification, a new result in Fig. 21 demonstrates a successful classification between two different damaged states (PDT8 and PDT17). This analysis confirms that the methodology is not inherently limited to binary classification between healthy and damaged states, but it is also effective in differentiating between distinct damage levels in a separate second stage. Specifically, the user may apply the methodology to multiple binary comparisons in parallel, or integrate them into a hierarchical or ensemble-based multi-class classification pipeline. Namely, the approach is adaptable and extensible to multi-class damage detection tasks when required.

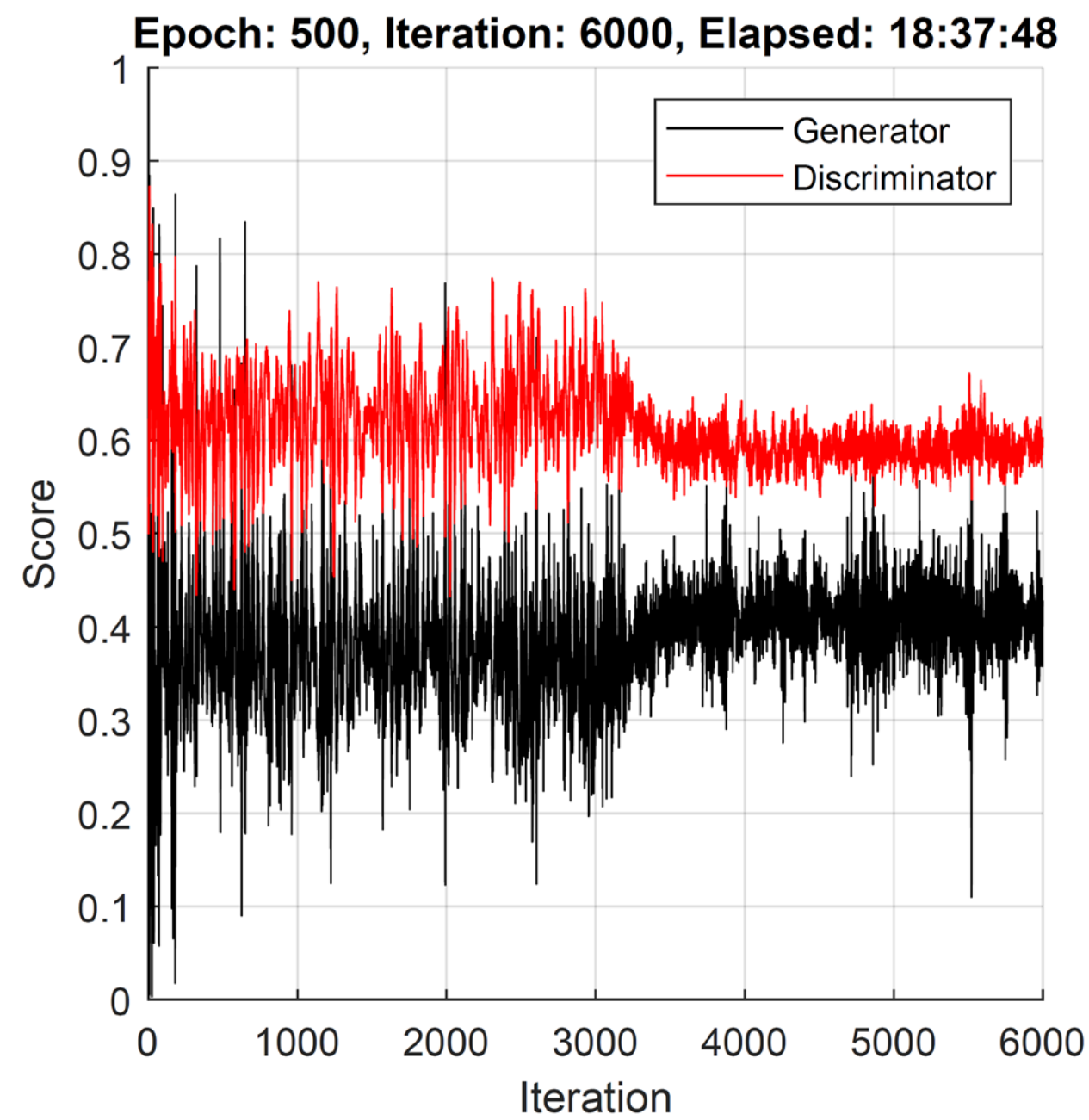

Fig. 17 Score for both networks when trained using damaged PDT17 (faked "Healthy") to damaged PDT17 of a different configuration measurements for 500$^{th}$ epochs

Generative adversarial networks are known for training instability and risks such as mode collapse due to poor hyper-parameter tuning or data-related issues as shown in Fig. 22 (left plot). A way to address this issue is by introducing small-scale noise to the training data, improving generalization as shown in Fig. 22 (right plot). Future extensions may explore architectural enhancements and regularization strategies (e.g., spectral normalization or Wasserstein loss) to further mitigate mode collapse and stabilize training.

Another concern is related to the fact that the support vector machine (Cortes and Vapnik 1995) classifier performance was not sufficient to distinguish damage from no damage. It was shown that classification is performed well, even when comparing healthy-to-healthy measurements. Interestingly though, the near-perfect classification may be used as an indication that the model learns something unrealistic as no new dynamics exists, and this justifies an unrealistic classification performance close to 100%.

Importantly, here, conditional generative adversarial network were used for generating structural vibration data conditioned on unknown damage states acting as a data-driven digital twin that mimics structural response under varying damage conditions. An immediate extension should be further searched in the use of reduce order modeling (Kuether et al. 2015; Roettgen et al. 2018; Vlachas et al. 2021, 2025; Bladh et al. 2001) to reduce the high computational cost.

Another approach to reduce the cost and increase accuracy should sought in the area of incorporating physics-based constraints in line with scientific machine learning (Qian et al. 2020; Sharma et al. 2024; Cuomo et al. 2022; Psaros et al. 2023; Bahmani and Sun 2024), including the incorporation of uncertainty (Kamariotis et al. 2025; Olivier et al. 2020; dos Santos et al. 2025; Lopez et al. 2025; Patelli et al. 2015). Physics can be incorporated in multiple ways, such as by using finite element model samples (Tsialiamanis et al. 2021), by using physical constraints derived from the governing equation of linear dynamic systems (Ge and Sadhu 2024), or by using probabilistic surrogate models (Mücke et al. 2023) with various engineering applications (Yan et al. 2022; Megia et al. 2024; Mousavi et al. 2025; Zhai et al. 2025). Nonetheless, the generative adversarial networks are inherently reduced order models as they learn latent, low-dimensional representations of high-dimensional structural dynamics, as a surrogate digital twins for reduces order models in a purely data-driven context.

So far, the method manages to provide a reliable prediction for damage. However, for other applications, such as if one wanted to predict erroneous measurements for vibration accelerometers, multiple fault measurement classes would be required to comprehensively consider normal, missing, minor error, outlier, square, trend, and/or drift data (Tang et al. 2019; Zhu et al. 2025; Gong etal. 2025; Liu et al. 2022). With regards to this point, future research is recommended for application on those cases. This would investigate the number of the potential inaccuracy types which results in the method to fail, and how the number of candidate inaccuracies effects the class prediction success. The reason lies into the fact that the number of inaccuracies would be prone to proliferation in a way that could potentially be detrimental to prediction performance.

Regarding the concern of the number 615 for samples shown in the confusion matrix, it reflects the result of > 15% holdout test split applied to the constructed dataset containing both real and model-generated signals. Specifically, the dataset used for classification comprises a total of 4100 signals: 1576 real signals (788 healthy + 788 damaged) and 2000 model-generated signals (1000 healthy + 1000 damaged). This composition ensures a balanced representation across both classes (healthy vs. damaged) and sources (real vs. generated). The classification task was conducted using a > 15% holdout split, yielding 615 test samples. To enhance clarity and reproducibility, a detailed summary of the sample composition is provided in Table 5. In supervised learning settings such as the support vector machine classifier

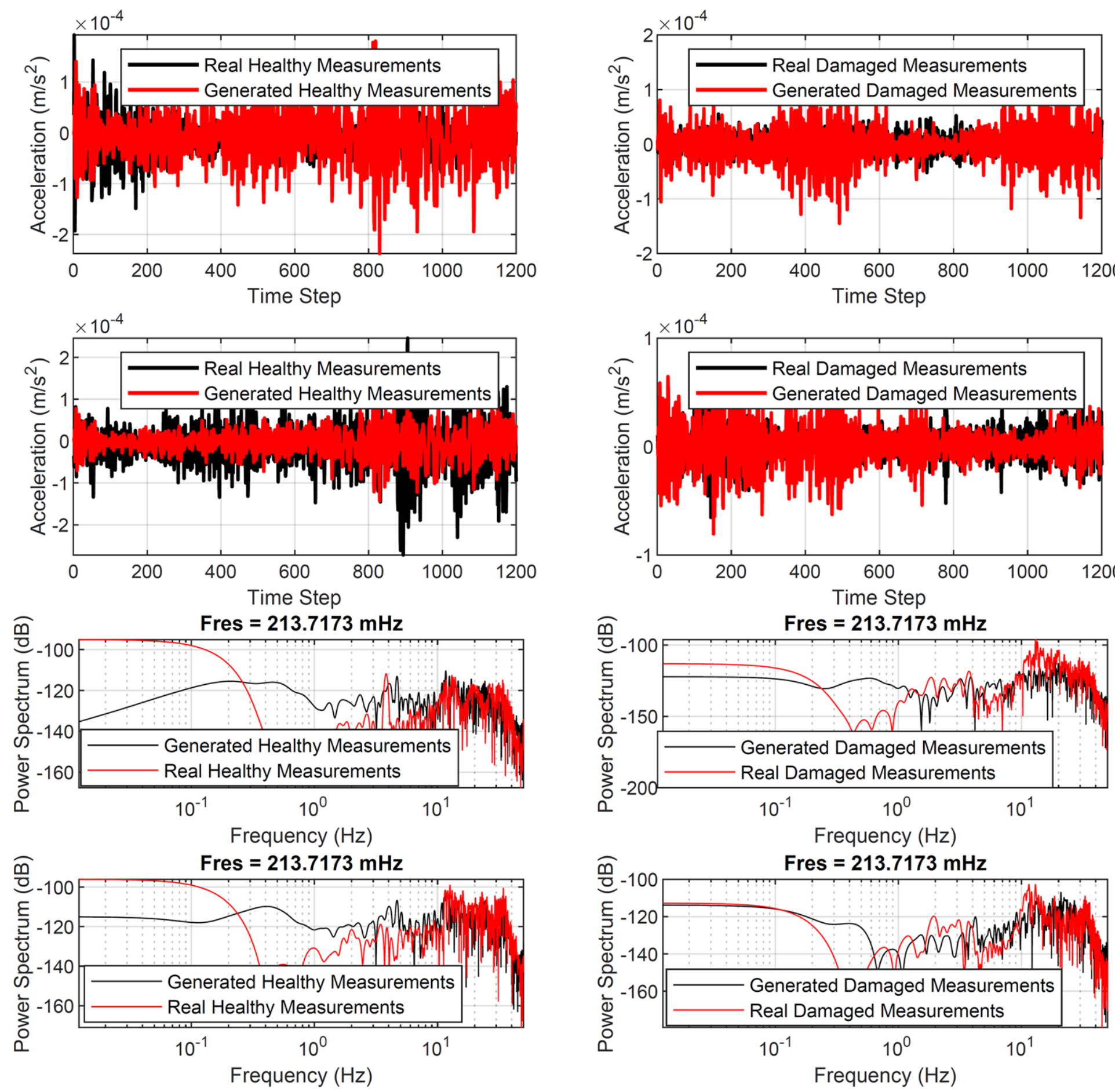

**Fig. 18** Time- and frequency-domain generated measurements for damaged PDT17 (faked "Healthy") to damaged PDT17 of a different configuration measurements for 500$^{th}$ epochs

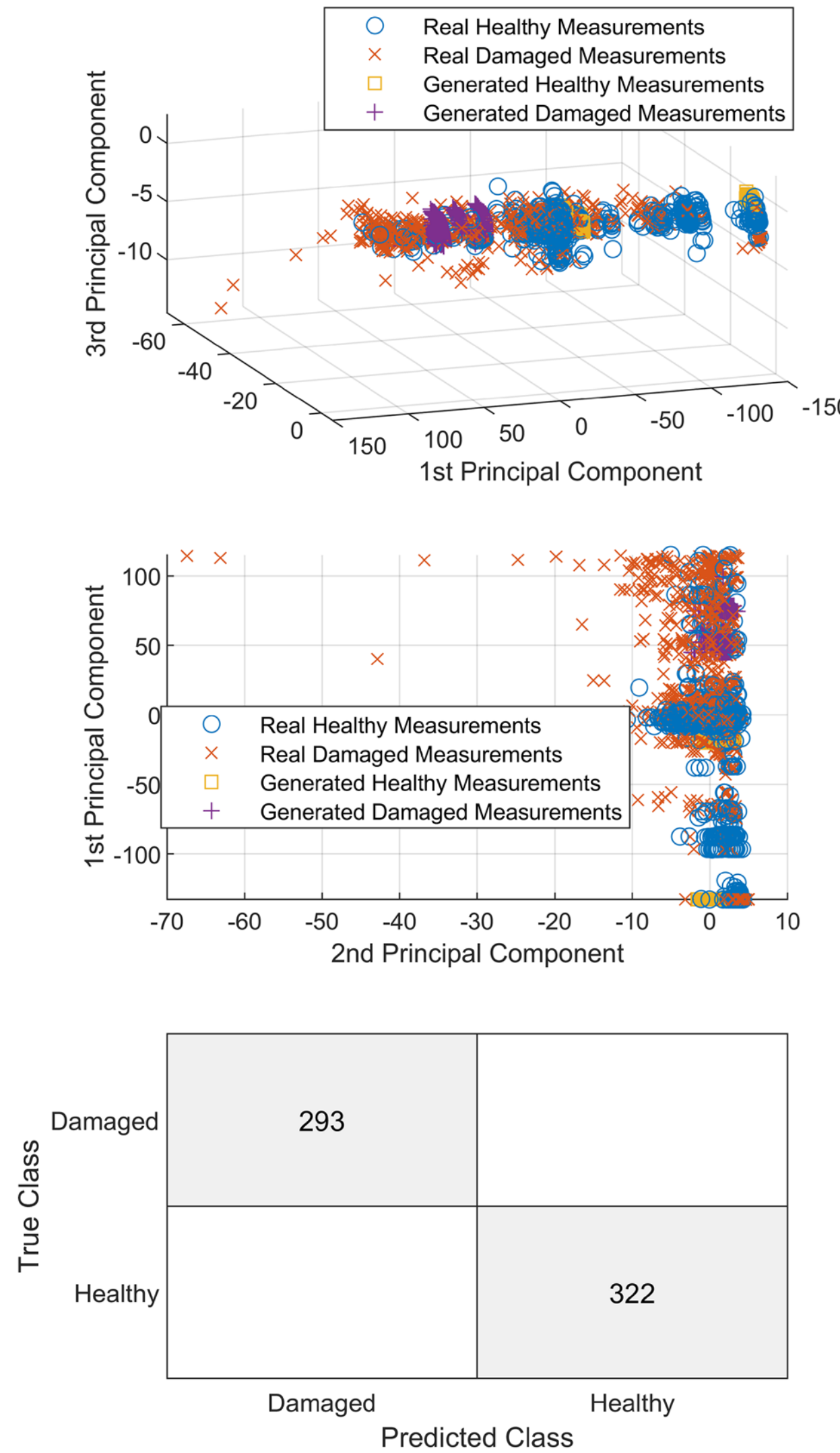

**Fig. 19** Real and generated measurements damaged PDT17 (faked "Healthy") to damaged PDT17 of a different configuration measurements on the first three principal components and support vector machine classifier performance for $500^{th}$ epochs

adopted in this study, the application of a holdout split is essential to objectively evaluate generalization performance. Without this partition, the classifier would be assessed on the same data it was trained on, leading to overly optimistic and potentially misleading performance metrics. Another application of > 50% holdout split is also included with the same conclusion.

Additional analyses in Fig. 23 illustrate the performance of classical damage detection approaches based on conventional features. Specifically, standard classifiers such as support vector machines is employed, as well as a new PCA angle showing the different damage state more clearly. While classical approaches can provide useful classification results in well-controlled settings, they are often limited by their dependence on handcrafted features and their inability to adapt to new or unseen damage patterns, especially under data-scarce conditions. In contrast, the proposed framework is unsupervised and simultaneously provides data augmentation through realistic signal generation.

Regarding the network algorithm parameters, the examinations so far showed a recommendation of as high as possible values for the filter size and the number of neurons in the convolutional layers. The higher the number of epochs and iterations also resulted in an improved performance of on the measurement generation. This is though in a contrast to the score representation where divergences are shown after a large number of iterations instead of continuous convergence to a stable dynamics of the two networks. However, the previous recommendations may sound restrictive or suboptimal since they lead to higher weights for back-propagation, or to a general ultimate higher computational cost. Despite this, the computational cost of this approach is bearable. This is attributed to three main reasons: the one-dimensional nature of the data, the low-signal training approach which may be implemented, and the potential use of high-performance graphics processing unit technologies. Last but not least, the training results and accuracy shows the normal variability of the neural networks training; due to the non-deterministic behavior of training, the model might differ slightly at every execution. The generator and discriminator architectures used in the proposed framework were designed to balance expressive capacity and computational tractability, given in Table 6.

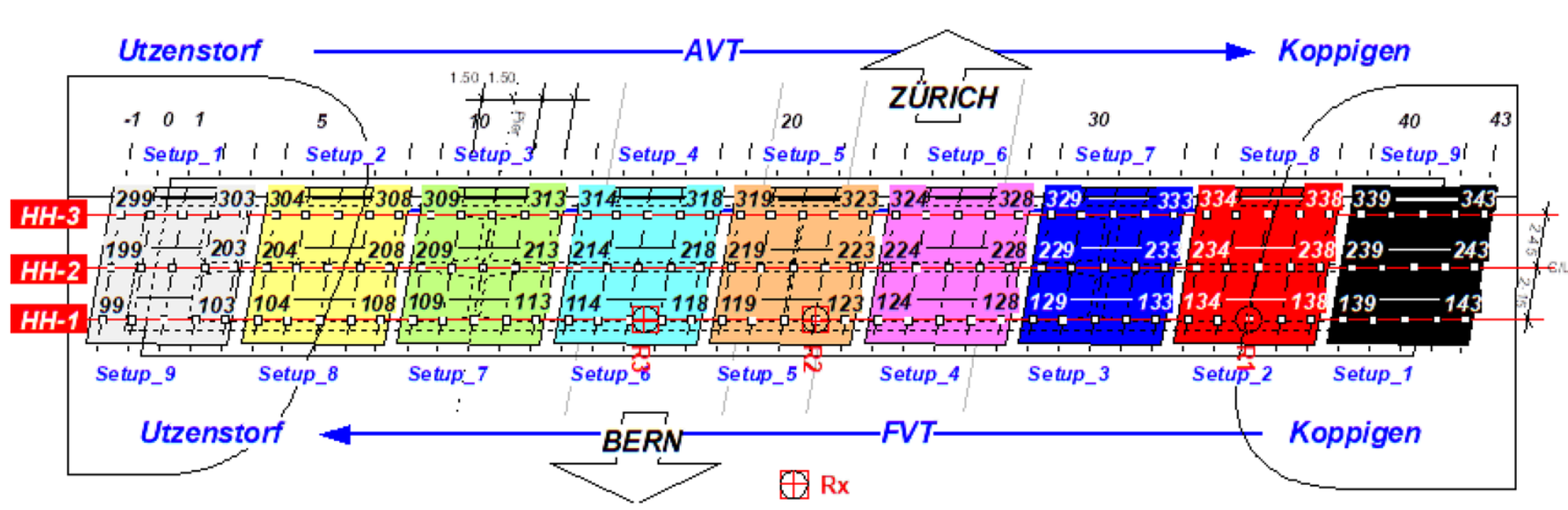

**Fig. 20** Girder, Top View of the Z24 Bridge for the sensor location (Peeters and De Roeck 2001) (research permission by KU Leuven)

**Table 4** Summary of Z24 dataset segmentation and preprocessing for model training and evaluation

| Parameter | Description |
|---|---|
| Measurement types | Ambient vibration tests (PDT01, PDT08, PDT17) |
| Sensor setup | 33 accelerometers |
| Sampling rate | 100 Hz |
| Raw signal length | 65,536 time steps per signal |
| Segmentation | 787 signals of 1201 time steps |
| Windowing | Non-overlapping windows |
| Preprocessing | Detrending, mean removal, standardization |
| Generated samples | 1000 signals per class |

A final concern is related to other types of neural network architectures such as the long short-term memory ones, and research is needed to further optimize the network for improved performance (Cao et al. 2025).

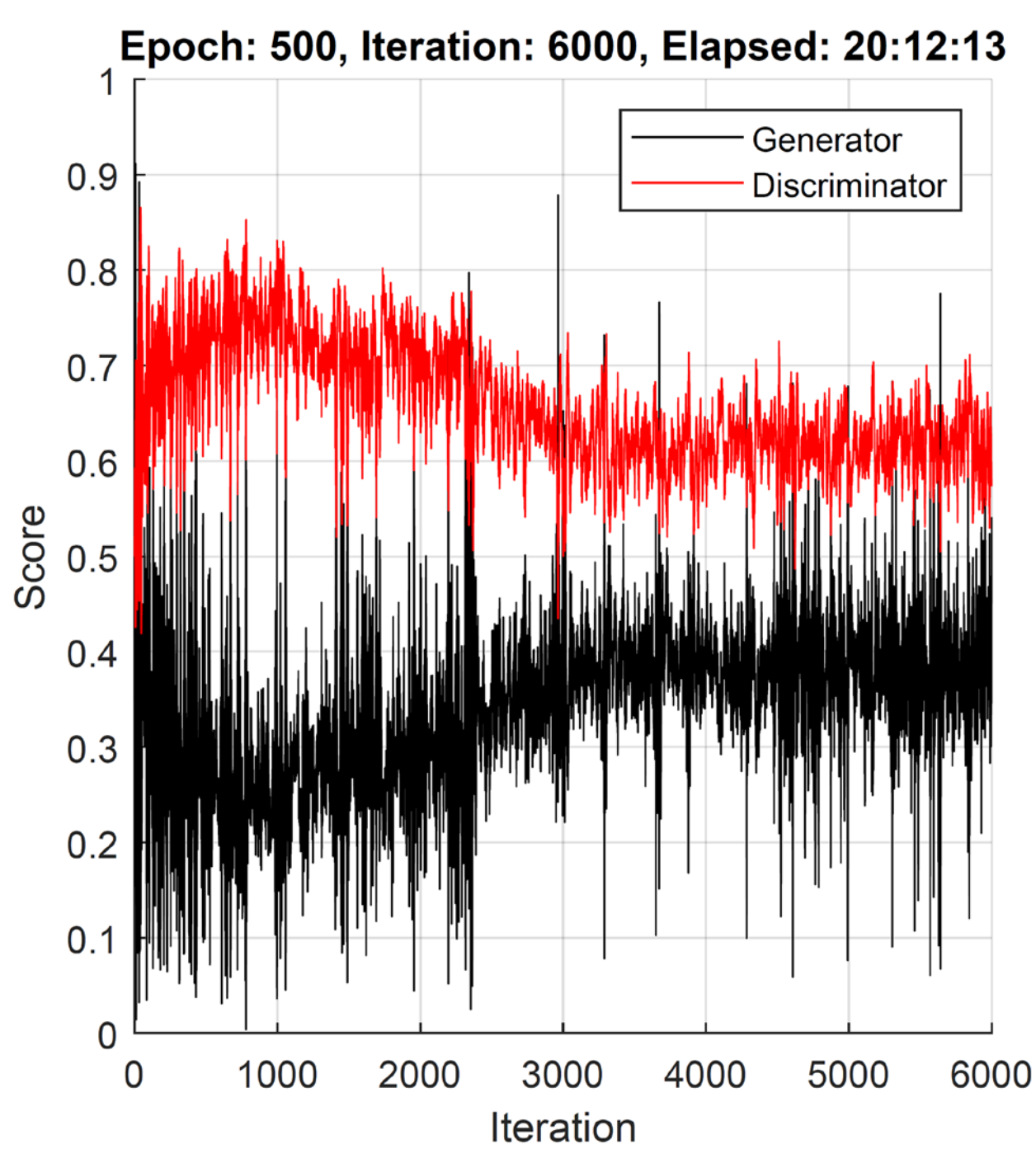

**Fig. 21** Score for both networks when trained using damage PDT8 to damage PDT17 measurements for $500^{th}$ epochs

## 6 Conclusion

This study presented a novel conditional generative adversarial network approach for structural health monitoring and digital twinning, validated on real-world benchmark Z24 Bridge measurements. By training on both healthy and damaged structural states, the proposed model successfully learnt to generate synthetic vibration signals that capture the temporal and statistical characteristics of real-world measurements. These generated signals can be used to augment training datasets, and support anomaly detection in a purely unsupervised manner. The findings reveal that the convergence behavior of the generative adversarial model during training, specifically the time and pattern of score stabilization, can serve as an implicit indicator of damage severity or novelty. Faster convergence is consistently observed in cases where the structural state remains unchanged (e.g., healthy-to-healthy measurement comparisons), whereas slower convergence correlates with increased damage evolution, suggesting the model's sensitivity to underlying structural dynamics. Importantly, classification tasks based on generated data consistently achieved high accuracy, highlighting the model's capability to preserve meaningful physical features despite working in a purely generative regime. However, results also caution against overreliance on classification accuracy alone, as models can achieve near-perfect scores even in unchanged conditions due to overfitting or generation of non-physical dynamics. The proposed framework demonstrates strong potential for scalable, automated structural health monitoring, particularly in scenarios where labeled data are scarce or damage scenarios cannot be compared to a previously known structural condition.

Overall, the method allowed for digital twinning with:

1. No need for prior knowledge of damage and health state.
2. Global end-to-end structural assessment.
3. Low cost computation using one-dimensional measurements.
4. Multiple indicators of fault anomaly based on the training and classification performance.
5. Direct data augmentation for all damage states.
6. Independent to the system application.

Importantly, the work leads to further integration of deep generative models into structural monitoring systems, offering robust tools for simulation, diagnosis, and early damage warning in structural systems. However, while the proposed framework effectively captures structural response patterns under varying damage states, its performance is influenced by the representativeness and diversity of the training data. The current setup uses a limited number of experimental signals, which may restrict generalization to more complex systems. Additionally, although the method operates in an unsupervised fashion, its convergence behavior is still implicitly dependent on the quality and consistency of input signals, which can be affected by sensor noise or environmental variability. Finally, the computational cost of training models remains a practical challenge.

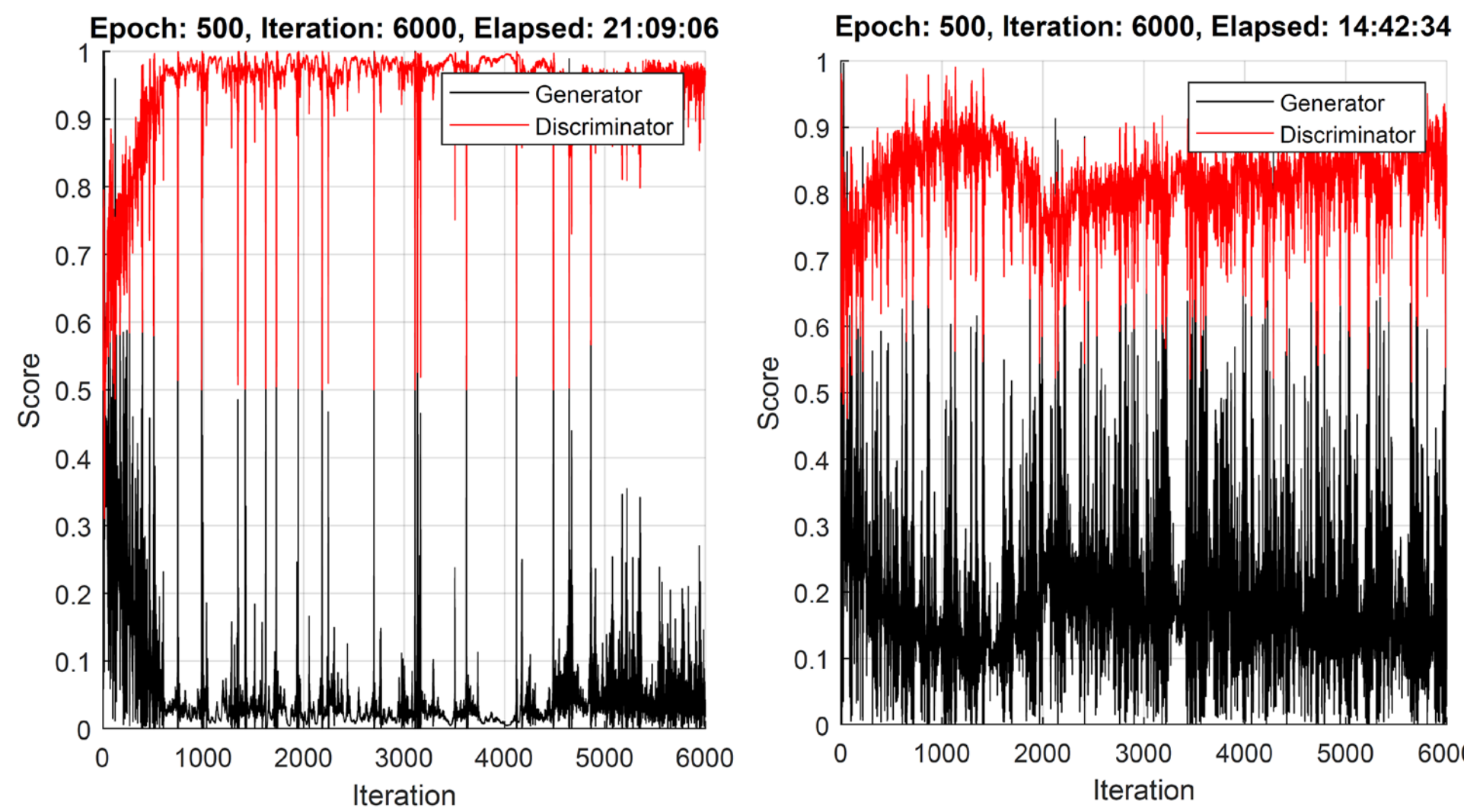

**Fig. 22** Training instability divergence of the model (left plot) and improved convergence after noise augmentation (right plot) when trained using healthy PDT1 to damage PDT8 measurements for $500^{th}$ epochs

**Table 5** Real and model-generated samples used for classification

| Condition | Source | Class Label | Samples |
|---|---|---|---|
| Healthy | Real | 1 (Real Healthy) | 787 |
| Damaged | Real | 2 (Real Damaged) | 788 |
| Healthy | Generated | 3 (GAN Healthy) | 1000 |
| Damaged | Generated | 4 (GAN Damaged) | 1000 |
| Total | All | 1–4 | 3575 |
| Test set (>15% split) | Subset | 1–4 | 615 |
| Test set (>50% split) | Subset | 1–4 | 1845 |

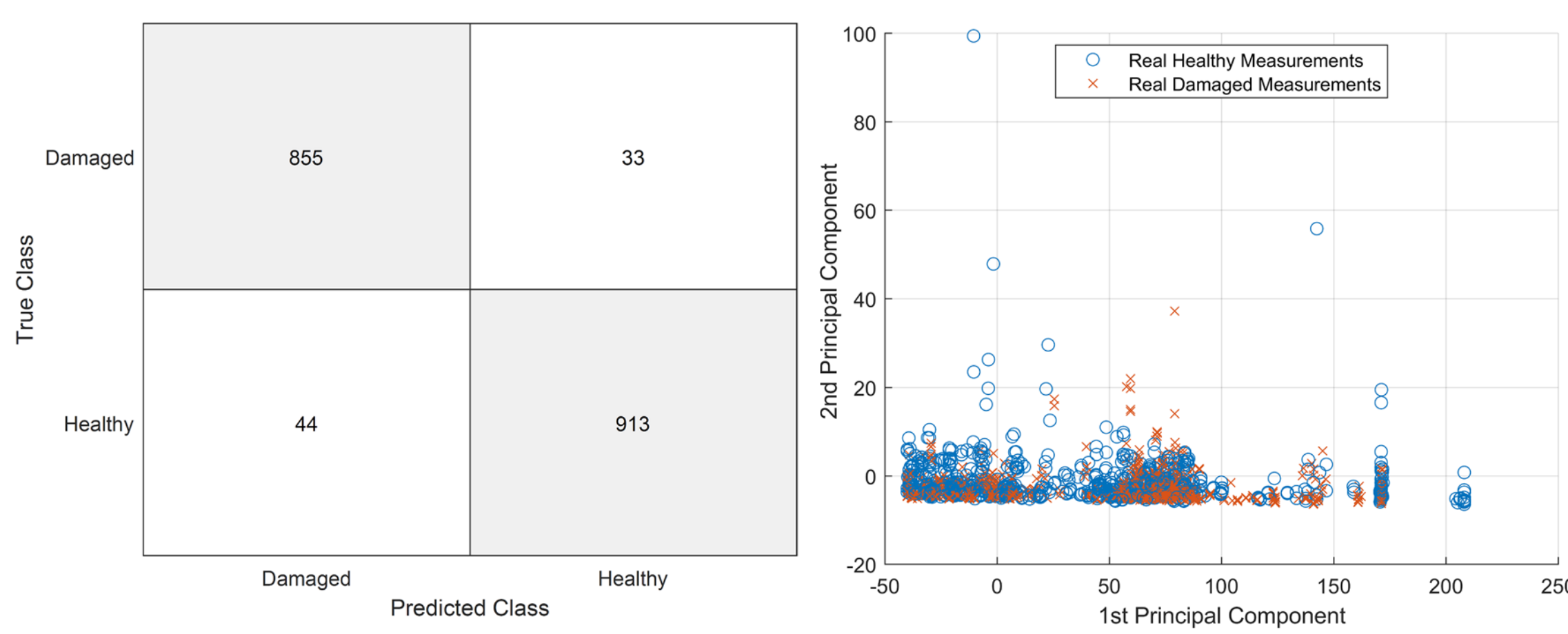

**Fig. 23** Support vector machine classifier performance for 1845 samples and real measurements healthy PDT1 to damage PDT8 measurements on a new principal component analysis angle

**Table 6** Key hyper-parameters and architectural choices

| Component | Final | Alternatives | Rationale |
|---|---|---|---|
| Latent dimension | 100 | 50, 128 | Balanced representation capacity |
| Learning rate | 0.0005 | 0.0002, 0.001 | 0.0005 gave stable convergence |
| Optimizer | Adam | – | Standard choice |
| Adam | 0.5/0.999 | 0.9/0.999 | Lower value improved stability |
| Epochs | 500 | 300, 1000 | Sufficient for convergence |
| Batch size | 128 | 64, 256 | Stable convergence |
| Filters (CNN) | 64 | 32, 128 | Good trade-off |
| Generator | T. Conv + BN + ReLU | – | Standard design |
| Discriminator | Conv + LReLU | – | Effective for fine signals |

**Acknowledgements** The authors gratefully acknowledge the KU Leuven Structural Mechanics Section, Ann Zwarts, and Pieter Reumers for providing the data.

**Author contributions** MI was responsible for conceptualization, methodology, software, validation, formal analysis, investigation, data curation, writing—original draft, writing—reviewing and editing, and visualization. ENP was involved in conceptualization, methodology, writing—reviewing and editing, and visualization.

**Funding** The University of Bath is acknowledged for providing the means to carry out the research, as well as for supporting open access publication through the Joint Information Systems Committee Read and Publish agreement. The authors received no external financial support for the research or authorship of this article.

**Data availability** The Z24 Bridge dataset is available from KU Leuven (Source: https://bwk.kuleuven.be/bwm/z24).

**Code availability** In this work, the basic codes for the evidence theory and the numerical results presented can be provided upon reasonable request.

**Replication of result** All results can be reproduced with the details presented in this paper. Codes for the evidence theory and the numerical results presented can also be provided upon reasonable request.

## Declarations

**Conflict of interest** The authors declare no potential conflict of interest with respect to the research, authorship, and/or publication of this article.

**Ethical approval** Not applicable.

## References

Andriotis CP, Papakonstantinou KG (2019) Managing engineering systems with large state and action spaces through deep reinforcement learning. Reliability Eng Syst Safe 191:106483

Arcieri G, Hoelzl C, Schwery O, Straub D, Papakonstantinou KG, Chatzi E (2023) Bridging Pomdps and Bayesian decision making for robust maintenance planning under model uncertainty: an application to railway systems. Reliability Eng Syst Safe 239:109496

Ates GC, Gorguluarslan RM (2021) Two-stage convolutional encoder-decoder network to improve the performance and reliability of deep learning models for topology optimization. Struct Multidisc Optim 63(4):1927–1950

Azimi M, Pekcan G (2020) Structural health monitoring using extremely compressed data through deep learning. Comput Aided Civil Infrastruct Eng 35(6):597–614

Azimi M, Eslamlou AD, Pekcan G (2020) Data-driven structural health monitoring and damage detection through deep learning: state-of-the-art review. Sensors 20(10):2778

Bahmani B, Sun W (2024) Physics-constrained symbolic model discovery for polyconvex incompressible hyperelastic materials. Int J Numer Meth Eng 125(15):e7473

Bao Y, Tang Z, Li H, Zhang Y (2019) Computer vision and deep learning-based data anomaly detection method for structural health monitoring. Struct Health Monit 18(2):401–421

Beck JL (2010) Bayesian system identification based on probability logic. Struct Control Health Monit 17(7):825–847

Behara R K, Saha A K (2024) Analysis of wind characteristics for grid-tied wind turbine generator using incremental generative adversarial network model. *IEEE Access*

Bernagozzi G, Mukhopadhyay S, Betti R, Landi L, Diotallevi PP (2022) Proportional flexibility-based damage detection for buildings in unknown mass scenarios: the case of severely truncated modal spaces. Eng Struct 259:114145

Bhattacharya A, Papakonstantinou KG, Warn GP, McPhillips L, Bilec MM, Forest CE, Hasan R, Chavda D (2025) Optimal life-cycle

adaptation of coastal infrastructure under climate change. Nat Commun 16(1):1076

Bilgin N, Olivier A (2025) Joint bayesian estimation of process and measurement noise statistics in nonlinear kalman filtering. Mech Syst Signal Process 223:111836

Bladh R, Castanier M, Pierre C (2001) Component-mode-based reduced order modeling techniques for mistuned bladed disks-part i: theoretical models. J Eng Gas Turbines Power 123(1):89–99

Bruneau M, Chang SE, Eguchi RT, Lee GC, O'Rourke TD, Reinhorn AM, Shinozuka M, Tierney K, Wallace WA, Von Winterfeldt D (2003) A framework to quantitatively assess and enhance the seismic resilience of communities. Earthq Spectra 19(4):733–752

Cao Q, Liu S, Varghese A J, Darbon J, Triantafyllou M, Karniadakis G E(2025) Automatic selection of the best neural architecture for time series forecasting via multi-objective optimization and pareto optimality conditions. *arXiv preprint* arXiv:2501.12215,

Catbas FN, Susoy M, Frangopol DM (2008) Structural health monitoring and reliability estimation: long span truss bridge application with environmental monitoring data. Eng Struct 30(9):2347–2359

Cha Y-J, Ali R, Lewis J, Buyukozturk O (2024) Deep learning-based structural health monitoring. Autom Constr 161:105328

Chua YK, Coble D, Razmarashooli A, Paul S, Martinez DAS, Hu C, Downey AR, Laflamme S (2025) Probabilistic machine learning pipeline using topological descriptors for real-time state estimation of high-rate dynamic systems. Mech Syst Signal Process 227:112319

Cimellaro GP, Reinhorn AM, Bruneau M (2010) Framework for analytical quantification of disaster resilience. Eng Struct 32(11):3639–3649

Cortes C, Vapnik V (1995) Support-vector networks. Machine Learning 20(3):273–297

Cuomo S, Di Cola VS, Giampaolo F, Rozza G, Raissi M, Piccialli F (2022) Scientific machine learning through physics-informed neural networks: where we are and what's next. J Sci Comput 92(3):88

Dang HV, Tran-Ngoc H, Nguyen TV, Bui-Tien T, De Roeck G, Nguyen HX (2020) Data-driven structural health monitoring using feature fusion and hybrid deep learning. IEEE Trans Autom Sci Eng 18(4):2087–2103

Dasgupta A, Patel DV, Ray D, Johnson EA, Oberai AA (2024) A dimension-reduced variational approach for solving physics-based inverse problems using generative adversarial network priors and normalizing flows. Comput Methods Appl Mech Eng 420:116682

Deraemaeker A, Worden K (2018) A comparison of linear approaches to filter out environmental effects in structural health monitoring. Mech Syst Signal Process 105:1–15

Deraemaeker A, Reynders E, De Roeck G, Kullaa J (2008) Vibration-based structural health monitoring using output-only measurements under changing environment. Mech Syst Signal Process 22(1):34–56

dos Santos KR, Chassignet AG, Pantoja-Rosero BG, Rezaie A, Savary OJ, Beyer K (2025) Uncertainty quantification for a deep learning models for image-based crack segmentation. J Civ Struct Heal Monit 15(4):1231–1269

Dunphy K, Sadhu A, Wang J (2022) Multiclass damage detection in concrete structures using a transfer learning-based generative adversarial networks. Struct Control Health Monit 29(11):e3079

Erazo K, Sen D, Nagarajaiah S, Sun L (2019) Vibration-based structural health monitoring under changing environmental conditions using kalman filtering. Mech Syst Signal Process 117:1–15

Farrar CR, Worden K (2007) An introduction to structural health monitoring. Philosophical Trans R Soc A Math Phys Eng Sci 365(1851):303–315

Garibaldi L, Marchesiello S, Bonisoli E (2003) Identification and up-dating over the z24 benchmark. Mech Syst Signal Process 17(1):153–161

Ge L, Sadhu A (2024) Domain adaptation for structural health monitoring via physics-informed and self-attention-enhanced generative adversarial learning. Mech Syst Signal Process 211:111236

Gong X, Song X, Zhu Y, Feng L, Lu X, Cai CS (2025) A two-stage temperature-driven method for detection of fault sensors and abnormal stress using structural health monitoring data. Struct Health Monit. https://doi.org/10.1177/14759217251315834

Goodfellow I, Pouget-Abadie J, Mirza M, Xu B, Warde-Farley D, Ozair S, Courville A, Bengio Y (2014). Generative adversarial nets. *Advances in neural information processing systems*, 27,

Gres S, Dohler M, Dertimanis VK, Chatzi EN (2025) Subspace-based noise covariance estimation for kalman filter in virtual sensing applications. Mech Syst Signal Process 222:111772

Guo X, Liu X, Królczyk G, Sulowicz M, Glowacz A, Gardoni P, Li Z (2022) Damage detection for conveyor belt surface based on conditional cycle generative adversarial network. Sensors 22(9):3485

Huang Y, Beck JL, Li H (2017) Bayesian system identification based on hierarchical sparse bayesian learning and gibbs sampling with application to structural damage assessment. Comput Methods Appl Mech Eng 318:382–411

Impraimakis M (2024) A convolutional neural network deep learning method for model class selection. Earthq Eng Struct Dynamics 53(2):784–814

Impraimakis M (2024) A Kullback–Leibler divergence method for input-system-state identification. J Sound Vibration 569:117965

Impraimakis M (2025) Deep recurrent-convolutional neural network learning and physics kalman filtering comparison in dynamic load identification. Struct Health Monit 24(3):1752–1782

Impraimakis M, Smyth AW (2022) Input-parameter-state estimation of limited information wind-excited systems using a sequential kalman filter. Struct Control Health Monitor 29(4):e2919

Impraimakis M, Smyth AW (2022) Integration, identification, and assessment of generalized damped systems using an online algorithm. J Sound Vibration 523:116696

Impraimakis M, Smyth AW (2022) A new residual-based kalman filter for real time input-parameter-state estimation using limited output information. Mech Syst Signal Process 178:109284

Impraimakis M, Smyth AW (2022) An unscented kalman filter method for real time input-parameter-state estimation. Mech Syst Signal Process 162:108026

Jacobs WR, Baldacchino T, Dodd T, Anderson SR (2018) Sparse bayesian nonlinear system identification using variational inference. IEEE Trans Autom Control 63(12):4172–4187

Kamariotis A, Vlachas K, Ntertimanis V, Koune I, Cicirello A, Chatzi E (2025) On the consistent classification and treatment of uncertainties in structural health monitoring applications. ASCE-ASME J Risk Uncertain Eng Syst B 11(1):011108

Kaya Y, Safak E (2015) Real-time analysis and interpretation of continuous data from structural health monitoring (shm) systems. Bull Earthq Eng 13:917–934

Keshmiry A, Hassani S, Mousavi M, Dackermann U (2023) Effects of environmental and operational conditions on structural health monitoring and non-destructive testing: a systematic review. Buildings 13(4):918

Kijewski-Correa T, Kwon DK, Kareem A, Bentz A, Guo Y, Bobby S, Abdelrazaq A (2013) Smartsync: an integrated real-time structural health monitoring and structural identification system for tall buildings. J Struct Eng 139(10):1675–1687

Kim D-H, Feng MQ (2007) Real-time structural health monitoring using a novel fiber-optic accelerometer system. IEEE Sens J 7(4):536–543

Kim S, Jwa M, Lee S, Park S, Kang N (2022) Deep learning-based inverse design for engineering systems: multidisciplinary design optimization of automotive brakes. Struct Multidisc Optim 65(11):323

Kingma D P(2014). Adam: A method for stochastic optimization. *arXiv preprint* arXiv: 1412.6980,

Kontoroupi T, Smyth AW (2016) Online noise identification for joint state and parameter estimation of nonlinear systems. ASCE-ASME J Risk Uncertainty Eng Syst A 2(3):B4015006

Kuether RJ, Deaner BJ, Hollkamp JJ, Allen MS (2015) Evaluation of geometrically nonlinear reduced-order models with nonlinear normal modes. AIAA J 53(11):3273–3285

Lei X, Sun L, Xia Y (2021) Lost data reconstruction for structural health monitoring using deep convolutional generative adversarial networks. Struct Health Monit 20(4):2069–2087

Li J, Guo F, Chen H (2024) A study on urban block design strategies for improving pedestrian-level wind conditions: Cfd-based optimization and generative adversarial networks. Energy Buildings 304:113863

Li X, Han X, Yang J, Wang J, Han G (2024) Transfer learning-based generative adversarial network model for tropical cyclone wind speed reconstruction from SAR images. IEEE Trans Geosci Remote Sens 62:1–16

Li Z-D, He W-Y, Ren W-X (2024) Structural damage identification based on wasserstein generative adversarial network with gradient penalty and dynamic adversarial adaptation network. Mech Syst Signal Process 221:111754

Liu G, Niu Y, Zhao W, Duan Y, Shu J (2022) Data anomaly detection for structural health monitoring using a combination network of ganomaly and cnn

Liu X, Yu J, Gong L, Liu M, Xiang X (2024) A gcn-based adaptive generative adversarial network model for short-term wind speed scenario prediction. Energy 294:130931

Lopez RH, Santos KR, Miguel LFF (2025) An efficient approach for taking into account uncertainties in structural parameters in performance-based design optimization. Eng Struct 336:120382

Luleci F, Catbas FN, Avci O (2023) Cyclegan for undamaged-to-damaged domain translation for structural health monitoring and damage detection. Mech Syst Signal Process 197:110370

Luleci F, Catbas FN, Avci O (2023) Generative adversarial networks for labeled acceleration data augmentation for structural damage detection. J Civil Struct Health Monit 13(1):181–198

Luo Y, Guo X, Wang L-K, Zheng J-L, Liu J-L, Liao F-Y (2023) Unsupervised structural damage detection based on an improved generative adversarial network and cloud model. J Low Frequency Noise Vibration Active Control 42(3):1501–1518

Maeck J, De Roeck G (2003a) Damage assessment using vibration analysis on the z24-bridge. Mech Syst Signal Process 17(1):133–142

Maeck J, Roeck G (2003) Description of z24 benchmark. Mech Syst Signal Process 17(1):127–131

Maeda H, Kashiyama T, Sekimoto Y, Seto T, Omata H (2021) Generative adversarial network for road damage detection. Comput Aided Civil Infrastruct Eng 36(1):47–60

Malekloo A, Ozer E, AlHamaydeh M, Girolami M (2022) Machine learning and structural health monitoring overview with emerging technology and high-dimensional data source highlights. Struct Health Monit 21(4):1906–1955

Mao J, Wang H, Spencer BF Jr (2021) Toward data anomaly detection for automated structural health monitoring: exploiting generative adversarial nets and autoencoders. Struct Health Monit 20(4):1609–1626

Masri S, Sheng L, Caffrey J, Nigbor R, Wahbeh M, Abdel-Ghaffar A (2004) Application of a web-enabled real-time structural health monitoring system for civilinfrastructure systems. Smart Mater Struct 13(6):1269

Megia M, Melero FJ, Chiachio M, Chiachio J (2024) Generative adversarial networks for improved model training in the context of the digital twin. Struct Control Health Monit 2024(1):9997872

Mirza M, Osindero S (2014) Conditional generative adversarial nets. *arXiv preprint* arXiv:1411.1784,

Morato PG, Andriotis CP, Papakonstantinou KG, Rigo P (2023) Inference and dynamic decision-making for deteriorating systems with probabilistic dependencies through bayesian networks and deep reinforcement learning. Reliability Eng Syst Safety 235:109144

Mousavi V, Rashidi M, Ghazimoghadam S, Mohammadi M, Samali B, Devitt J (2025) Transformer-based time-series gan for data augmentation in bridge digital twins. Autom Constr 175:106208

Mücke NT, Pandey P, Jain S, Bohté SM, Oosterlee CW (2023) A probabilistic digital twin for leak localization in water distribution networks using generative deep learning. Sensors 23(13):6179

Okasha NM, Frangopol DM, Deco A (2010) Integration of structural health monitoring in life-cycle performance assessment of ship structures under uncertainty. Mar Struct 23(3):303–321

Olivier A, Giovanis DG, Aakash B, Chauhan M, Vandanapu L, Shields MD (2020) Uqpy: a general purpose python package and development environment for uncertainty quantification. J Comput Sci 47:101204

Patelli E, Alvarez DA, Broggi M, Angelis MD (2015) Uncertainty management in multidisciplinary design of critical safety systems. J Aerospace Inf Syst 12(1):140–169

Peeters B, Roeck G (2001) One-year monitoring of the z24-bridge: environmental effects versus damage events. Earthquake Eng Struct Dynamics 30(2):149–171

Prajapati KK, Ghosh A, Mitra M (2025) Semi-supervised generative adversarial network (sgan) for damage detection in a composite plate using guided wave responses. Mech Syst Signal Process 232:112686

Psaros AF, Meng X, Zou Z, Guo L, Karniadakis GE (2023) Uncertainty quantification in scientific machine learning: methods, metrics, and comparisons. J Comput Phys 477:111902

Qian C, Ye W (2021) Accelerating gradient-based topology optimization design with dual-model artificial neural networks. Struct Multidisc Optim 63(4):1687–1707

Qian E, Kramer B, Peherstorfer B, Willcox K (2020) Lift & learn: physics-informed machine learning for large-scale nonlinear dynamical systems. Physica D 406:132401

Rainieri C, Fabbrocino G, Cosenza E (2011) Near real-time tracking of dynamic properties for standalone structural health monitoring systems. Mech Syst Signal Process 25(8):3010–3026

Ramu P, Thananjayan P, Acar E, Bayrak G, Park JW, Lee I (2022) A survey of machine learning techniques in structural and multidisciplinary optimization. Struct Multidisc Optim 65(9):266

Rastin Z, Ghodrati Amiri G, Darvishan E (2021) Generative adversarial network for damage identification in civil structures. Shock and Vibration 2021(1):3987835

Razmarashooli A, Chua YK, Barzegar V, Salazar D, Laflamme S, Hu C, Downey AR, Dodson J, Schrader PT (2025) Real-time state estimation of nonstationary systems through dominant fundamental frequency using topological data analysis features. Mech Syst Signal Process 224:112048

Roettgen DR, Seeger B, Tai WC, Baek S, Dossogne T, Allen MS, Kuether RJ, Brake MR, Mayes RL (2018) A comparison of reduced order modeling techniques used in dynamic substructuring, The Mechanics of Jointed Structures: Recent Research and Open Challenges for Developing Predictive Models for Structural Dynamics. Springer, Cham

Roveri N, Severa L, Milana S, Tronci EM, Culla A, Betti R, Carcaterra A (2025) Adacta-advanced component analysis technique for damage detection. Struct Health Monit. https://doi.org/10.1177/14759217251326599

Sajeeda A, Hossain BM (2022) Exploring generative adversarial networks and adversarial training. Int J Cognitive Comput Eng 3:78–89

Seventekidis P, Giagopoulos D, Arailopoulos A, Markogiannaki O (2020) Structural health monitoring using deep learning with optimal finite element model generated data. Mech Syst Signal Process 145:106972

Sharma H, Novak L, Shields M (2024) Physics-constrained polynomial chaos expansion for scientific machine learning and uncertainty quantification. Comput Methods Appl Mech Eng 431:117314

Shim S, Kim J, Lee S-W, Cho G-C (2022) Road damage detection using super-resolution and semi-supervised learning with generative adversarial network. Autom Constr 135:104139

Smarsly K, Hartmann D, Law KH (2013) A computational framework for life-cycle management of wind turbines incorporating structural health monitoring. Struct Health Monit 12(4):359–376

Sohn H (2007) Effects of environmental and operational variability on structural health monitoring. Philosophical Trans R Soc A Math Phys Eng Sci 365(1851):539–560

Soleimani-Babakamali MH, Soleimani-Babakamali R, Nasrollahzadeh K, Avci O, Kiranyaz S, Taciroglu E (2023) Zero-shot transfer learning for structural health monitoring using generative adversarial networks and spectral mapping. Mech Syst Signal Process 198:110404

Tan RK, Zhang NL, Ye W (2020) A deep learning-based method for the design of microstructural materials. Struct Multidisc Optim 61(4):1417–1438

Tang Z, Chen Z, Bao Y, Li H (2019) Convolutional neural network-based data anomaly detection method using multiple information for structural health monitoring. Struct Control Health Monit 26(1):e2296

Teughels A, De Roeck G (2004) Structural damage identification of the highway bridge z24 by fe model updating. J Sound Vib 278(3):589–610

Teymouri D, Sedehi O, Katafygiotis LS, Papadimitriou C (2023) Input-state-parameter-noise identification and virtual sensing in dynamical systems: a bayesian expectation-maximization (bem) perspective. Mech Syst Signal Process 185:109758

Thanh-Tung H, Tran T, Venkatesh S (2019). Improving generalization and stability of generative adversarial networks. *arXiv preprint* arXiv:1902.03984,

Thelen A, Zhang X, Fink O, Lu Y, Ghosh S, Youn BD, Todd MD, Mahadevan S, Hu C, Hu Z (2022) A comprehensive review of digital twin-part 1: modeling and twinning enabling technologies. Struct Multidisc Optim 65(12):354

Tilon S, Nex F, Kerle N, Vosselman G (2020) Post-disaster building damage detection from earth observation imagery using unsupervised and transferable anomaly detecting generative adversarial networks. Remote Sens 12(24):4193

Torti M, Venanzi I, Laflamme S, Ubertini F (2022) Life-cycle management cost analysis of transportation bridges equipped with seismic structural health monitoring systems. Struct Health Monit 21(1):100–117

Tsialiamanis G, Wagg DJ, Dervilis N, Worden K (2021) On generative models as the basis for digital twins. Data-Centric Eng 2:e11

Tsialiamanis G, Champneys M, Dervilis N, Wagg D, Worden K (2022) On the application of generative adversarial networks for nonlinear modal analysis. Mech Syst Signal Process 166:108473

Velde M, Anastasopoulos D, Backer H, Reynders E, Lombaert G (2025) Modal strain monitoring of a post-tensioned concrete girder bridge: influence of temperature and solar irradiation. Eng Struct 329:119832

Vlachas K, Tatsis K, Agathos K, Brink AR, Chatzi E (2021) A local basis approximation approach for nonlinear parametric model order reduction. J Sound Vib 502:116055

Vlachas K, Simpson T, Garland A, Quinn DD, Farhat C, Chatzi E (2025) Reduced order modeling conditioned on monitored features for response and error bounds estimation in engineered systems. Mech Syst Signal Process 226:112261

Wagg D, Worden K, Barthorpe R, Gardner P (2020) Digital twins: state-of-the-art and future directions for modeling and simulation in engineering dynamics applications. ASCE-ASME J Risk Uncertainty Eng Syst B 6(3):030901

Wiatrak M, Albrecht S V, Nystrom A (2019). Stabilizing generative adversarial networks: A survey. *arXiv preprint* arXiv: 1910.00927,

Xiao Z, Rao J, Eisenträger S, Yuen K-V, Hadigheh SA (2025) Generative adversarial network-based ultrasonic full waveform inversion for high-density polyethylene structures. Mech Syst Signal Process 224:112160

Yan K, Chong A, Mo Y (2020) Generative adversarial network for fault detection diagnosis of chillers. Build Environ 172:106698

Yan K, Chen X, Zhou X, Yan Z, Ma J (2022) Physical model informed fault detection and diagnosis of air handling units based on transformer generative adversarial network. IEEE Trans Industr Inf 19(2):2192–2199

Ye L, Peng Y, Li Y, Li Z (2024) A novel informer-time-series generative adversarial networks for day-ahead scenario generation of wind power. Appl Energy 364:123182

Yonekura K, Miyamoto N, Suzuki K (2022) Inverse airfoil design method for generating varieties of smooth airfoils using conditional wgan-gp. Struct Multidisc Optim 65(6):173

Yu Y, Hur T, Jung J, Jang IG (2019) Deep learning for determining a near-optimal topological design without any iteration. Struct Multidisc Optim 59(3):787–799

Yuen K-V, Liu Y-S, Yan W-J (2022) Estimation of time-varying noise parameters for unscented kalman filter. Mech Syst Signal Process 180:109439

Zhai G, Xu Y, Spencer BF (2025) Bidirectional graphics-based digital twin framework for quantifying seismic damage of structures using deep learning networks. Struct Health Monit 24(1):86–110

Zhang C, Liu L, Wang H, Song X, Tao D (2022) Scgan: stacking-based generative adversarial networks for multi-fidelity surrogate modeling. Struct Multidisc Optim 65(6):163

Zhang X, Li D, Fu X (2024) A novel wasserstein generative adversarial network for stochastic wind power output scenario generation. IET Renew Power Gener 18(16):3731–3742

Zhong J, Huyan J, Zhang W, Cheng H, Zhang J, Tong Z, Jiang X, Huang B (2023) A deeper generative adversarial network for grooved cement concrete pavement crack detection. Eng Appl Artif Intell 119:105808

Zhu Y-K, Wan H-P, Todd MD (2025) A method for reconstruction of structural health monitoring data using wgangp with u-net generator. Struct Health Monit. https://doi.org/10.1177/14759217241313438